\pdfoutput=1

\documentclass[11pt]{article}

\usepackage[]{acl}

\usepackage{times}
\usepackage{latexsym}

\usepackage[T1]{fontenc}

\usepackage[utf8]{inputenc}

\usepackage{microtype}

\usepackage{inconsolata}

\usepackage{times}
\usepackage{latexsym}
\usepackage{tcolorbox}
\usepackage{multirow}
\usepackage{tikz}
\usepackage{listings}
\usepackage{capt-of}
\usepackage{graphicx}  
\usepackage{pgfplots}
\pgfplotsset{compat=1.12}
\usepackage{amsmath}
\usepackage{multicol}
\usepackage{booktabs}
\usepackage{colortbl,array,xcolor}
\usepackage{xspace}

\newcommand{\rib}{$\mathbb{BERRI}$\xspace}
\newcommand{\model}{$\mathbb{TART}$\xspace}
\newcommand{\task}{$\mathbb{X}^2$-Retrieval\xspace}

\usetikzlibrary{intersections}

\usepackage{caption}
\usepackage{subcaption}
\usepackage{graphicx}
\usepackage{amsfonts}
\usepackage{booktabs}
\usepackage{xcolor}
\usepackage{soul}
\usepackage{graphicx}

\newcommand{\hlc}[2][yellow]{{%
    \colorlet{foo}{#1}%
    \sethlcolor{foo}\hl{#2}}%
}

\definecolor{Gray}{gray}{0.85}
\definecolor{LightCyan}{rgb}{0.88,1,1}

\newcolumntype{a}{>{\columncolor{Gray}}c}
\newcolumntype{b}{>{\columncolor{LightCyan}}c}

\usetikzlibrary{shapes.geometric}
\usepackage{xspace}

\usepackage{multirow} 
\usepackage{amsmath}

\usepackage{pifont}
\newcommand{\xmark}{\ding{55}}
\usepackage{algorithm}
\usepackage{algpseudocode}%

\title{Task-aware Retrieval with Instructions}
\author{\parbox{0.9\linewidth}{\centering{
Akari Asai$^{\dagger\ddagger}$, Timo Schick$^{\dagger}$, Patrick Lewis$^{\dagger}$, Xilun Chen$^{\dagger}$, Gautier Izacard$^{\dagger\spadesuit}$, Sebastian Riedel$^{\clubsuit}$, Hannaneh Hajishirzi$^{\ddagger\heartsuit}$, Wen-tau Yih$^{\dagger}$} \\
{\rm $^\dagger$Meta AI ~~$^\ddagger$University of Washington~~$^\spadesuit$ ENS, PSL University \& Inria \\$^\heartsuit$Allen Institute for AI~~$^\clubsuit$University College London} \\
}
}

\begin{document}
\maketitle
\begin{abstract}

We study the problem of \emph{retrieval with instructions}, where users of a retrieval system explicitly describe their intent along with their queries.
We aim to develop a general-purpose task-aware retrieval system using multi-task instruction tuning, which can follow human-written instructions to find the best documents for a given query.
{We introduce the first large-scale collection of approximately 40 retrieval datasets with instructions, \rib, and present \model, a multi-task retrieval system trained on \rib with instructions. }
{\model shows strong capabilities to adapt to a new retrieval task via instructions and advances the state of the art on two zero-shot retrieval benchmarks, BEIR and LOTTE, outperforming models up to three times larger.}
We further introduce a new evaluation setup, \task to better reflect real-world scenarios, {where diverse domains and tasks are pooled and a system needs to find documents aligning users' intents}. In this setup, \model significantly outperforms competitive baselines, further demonstrating the effectiveness of guiding retrieval with instructions.\footnote{Code, data and pretrained model checkpoints are available at \url{https://github.com/facebookresearch/tart}. }

\end{abstract}

\section{Introduction}
Information retrieval (IR) is the task of finding {\it relevant} documents from a large collection of texts to fulfill a user's information need, typically expressed in the form of a textual query~\cite{singhal2001modern}.
The notion of relevance from the user's perspective (i.e., {\it intent}) can be amorphous~\cite{8160827}, and a query alone may not fully capture user information needs~\cite{ruthven_lalmas_2003,Taylor1962ThePO}.
\begin{figure}[t!]
\includegraphics[width=8cm]{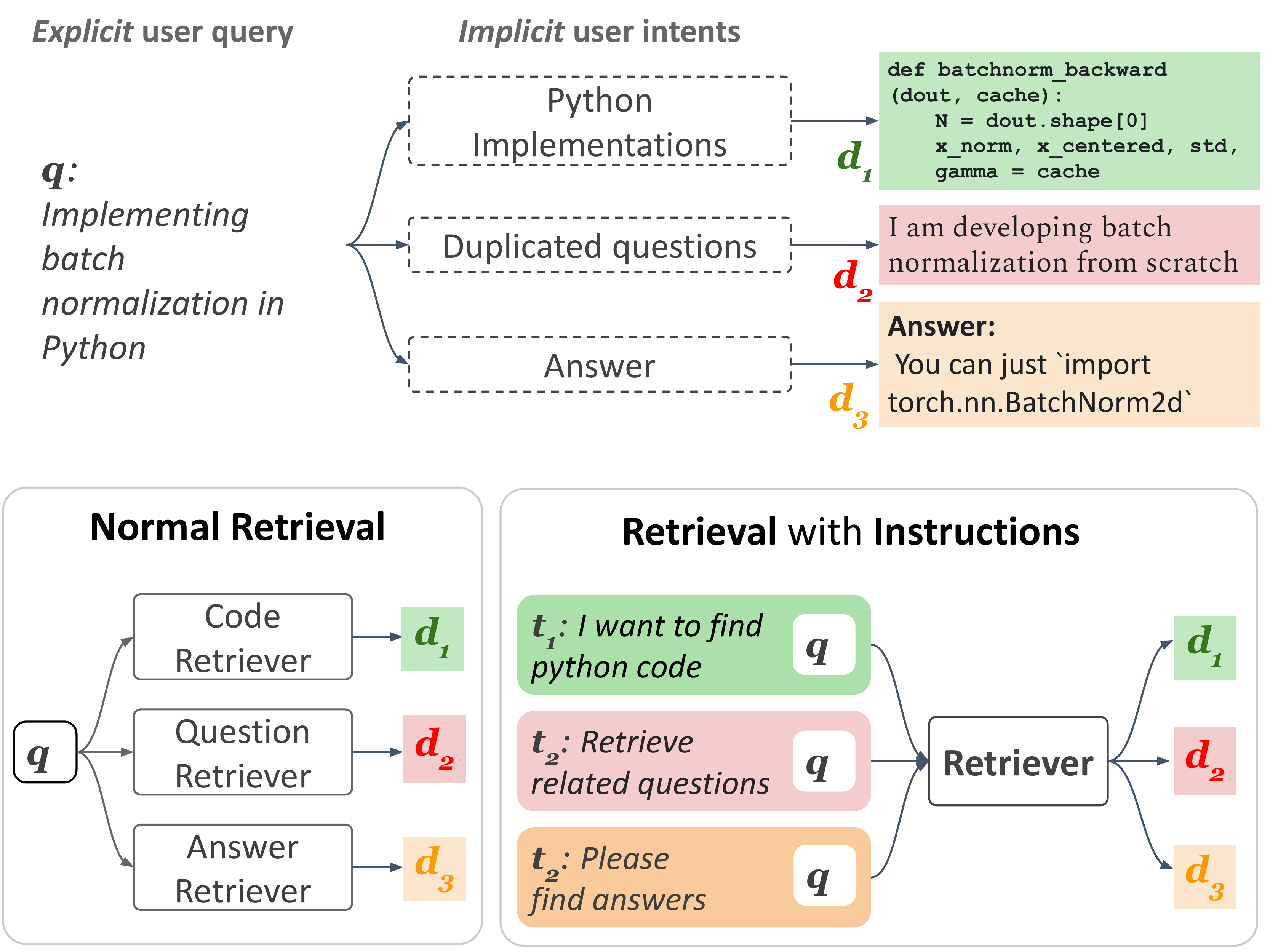}\caption{
User intents are not fully captured in query $\boldsymbol{q}$ only (top). 
Conventional approaches (bottom left) take a query and retrieve documents from a closed corpus using a task-specific retriever. 
\emph{Retrieval with instructions} (bottom right) takes a query and explicit intent and retrieves documents aligning with the user's expectations.
} \label{fig:teaser}
\end{figure}
As illustrated in Figure~\ref{fig:teaser} (top), given the same query, ``\textit{implementing batch normalization in Python},'' a user may want to retrieve a passage that describes how to do the task or to identify a similar query, or even to directly locate a code snippet.

Most existing work tries to learn those {\it implicit} intents from labeled data (e.g., pairs of queries and relevant documents), yielding separate models for different intents 
as shown in the bottom left of  Figure~\ref{fig:teaser}. 
This approach has several limitations. 
First, a vast number of annotated examples may be required to train a model to capture the task-specific notion of relevance, while they could benefit from the abundance of data available from related tasks.
Second, a model trained on one task may not easily transfer to new tasks that are not closely related. 

In this work we advocate for a new task formulation, \textit{retrieval with instructions}, to {\it explicitly} model a user's search intent {by providing a natural language description of the search task (i.e., an instruction). }
Here, the goal of retrieval systems is to retrieve documents that are both relevant to the query \emph{and} well-suited to the instructions.

{
Despite active research in other settings, instruction-following has not been systematically explored in retrieval, partly due to the lack of annotated resources.
}
To facilitate research in retrieval with instructions, we introduce \rib ({\bf B}ank of {\bf E}xplicit {\bf R}et{\bf R}ieval {\bf I}nstructions), a collection of approximately 40 retrieval datasets with diverse instructions in a unified format{, covering 10 diverse domains. }
{Each task has on average 3.5 diverse instructions annotated by experts, following our novel instruction schema for retrieval tasks.}

{
{
We use \rib to train \model (\textbf{T}ask-\textbf{a}ware \textbf{R}e\textbf{T}riever), a single multi-task retrieval system 
that follows instructions to perform diverse tasks with no parameter updates on each task. 
}}
We employ two widely explored architectures: 
\model-dual is a dense dual-encoder architecture, retrieving documents based on the similarity of independently encoded query and document embeddings; \model-full calculates probabilities of a document being relevant to the query according to the instruction using a cross-encoder. 
\model is trained with carefully designed negative samples, including our novel {instruction-unfollowing negatives samples}.

The \model models, particularly \model-full yields state-of-the-art results on two popular zero-shot retrieval benchmarks, BEIR~\cite{thakur2021beir} and LOTTE-pooled~\cite{santhanam-etal-2022-colbertv2}, outperforming systems using three times more parameters (\citealt{nogueira-etal-2020-document, ni2021large, muennighoff2022sgpt}) as well as task-specific retrievers trained on millions of automatically generated examples~\cite{promptagator,wang-etal-2022-gpl}.

We further introduce a new evaluation setup, \task ({\bf Cross}-task {\bf Cross}-domain Retrieval), where a system needs to handle queries with diverse intents to find relevant documents from a large-scale, cross-domain pooled corpus{, simulating challenges in real-world retrieval applications.}
In this under-explored setting, \model outperforms other state-of-the-art methods, demonstrating its ability to find documents in a large-scale open-domain corpus by leveraging explicit textual intents. 
Our analysis shows that training a model on diverse tasks with instructions, {our new negative samples leveraging instructions} and giving informative instructions are crucial.

In summary, our contributions are as follows: 
\vspace{-0.1cm}
\begin{itemize}
\itemsep-0.2em 
\item \emph{Retrieval with instructions}, a new formulation to model users' intent {\it explicitly}
(Section~\ref{sec:task_formulation}). 
\item {\rib, a new large-scale collection of approximately 40 retrieval datasets in diverse domains with instructions (Section~\ref{sec:dataaset}). }
\item {\model, a task-aware retriever trained on \rib that advances state of the art on zero-shot and cross-task retrieval (Section~\ref{sec:method})}. 
\end{itemize}

\section{Background and Related Work}
\label{sec:related_work}
\paragraph{Zero-shot training of retrievers.}
Recent neural retrievers~\cite{karpukhin-etal-2020-dense,lee-etal-2019-latent,colbert} show their superiority over term-based retrievers (e.g., BM25; \citealt{robertson2009probabilistic}) across domains when training data is abundant~\cite{luo2022improving,asai2021one,petroni-etal-2021-kilt}.
{Due to the high cost of annotating retrieval datasets for new target tasks, improving neural retrievers in zero-shot settings is an active area of study. 
The first line of work uses purely unsupervised approaches, such as pre-training neural retrievers on {unlabeled} data (e.g., Contriever; \citealt{izacard2022unsupervised}).
}
The second line of work trains a single retrieval system on large-scale supervised datasets such as MS MARCO~\cite{msmarco} and then performs a transfer to new datasets~\cite{izacard2022unsupervised,colbert,nogueira-etal-2020-document,chen2021salient}, which often struggle with tasks unlike those used for training~\cite{promptagator}. 
To address this, the third paradigm trains customized retrievers for each task using unlabeled corpora, leveraging another model to automatically generate training data~\cite{wang-etal-2022-gpl}. 
Concurrent to our work, \citet{promptagator} use task-specific templates and few-shot samples to automatically generate in-domain training queries given randomly sampled documents from the target corpus using FLAN~\cite{wei2022finetuned}. 
It often entails running massive LMs and training separate retrievers, resulting in slow and costly adaptation. 
\begin{table*}[t!]
\renewcommand{\arraystretch}{1.2}
\setlength{\tabcolsep}{2pt}
\footnotesize
    \centering
    \begin{tabular}{ll}
\toprule
\textbf{Dataset} & \textbf{Instruction} \\\midrule
NQ & \texttt{Retrieve a \hlc[cyan!30]{Wikipedia} \hlc[green!30]{paragraph} that \hlc[pink]{answers this question}.}  \\
Med Simple & \texttt{Your task is to find a \hlc[pink]{simplified} \hlc[green!30]{paragraph} of this paragraph from a \hlc[cyan!30]{medical paper.} }\\
QReCC & \texttt{Find a \hlc[green!30]{dialogue response} from  \hlc[cyan!30]{dialogue history} to \hlc[pink]{answer the user's question}. }  \\
Arguana & \texttt{Retrieve a \hlc[green!30]{paragraph} from \hlc[cyan!30]{an argument website} that \hlc[pink]{argues against the following argument}.}  \\
SciFact & \texttt{Find a \hlc[green!30]{sentence} from a \hlc[cyan!30]{scientific paper} to check if \hlc[pink]{the statement is correct or not}. } \\
MultiLexSum & \texttt{I want to find the \hlc[green!30]{one-sentence} \hlc[pink]{summary of this} \hlc[cyan!30]{legal case}.} \\
\bottomrule
 \end{tabular}
    \caption{Example instructions for natural questions (NQ; \citealt{47761}), medical text simplification (Med Simple;~\citealt{devaraj-etal-2021-paragraph}), QReCC~\cite{qrecc}, Arguana~\cite{wachsmuth-etal-2018-retrieval}, SciFact~\cite{wadden-etal-2020-fact} and MultiLexSum~\cite{shen2022multilexsum}. Each instruction defines \hlc[pink]{\emph{intent}}, \hlc[cyan!30]{\emph{domain}} and \hlc[green!30]{\emph{unit}}. 
    }\label{tab:instructions}
\end{table*}
\paragraph{Instruction tuning.}
Training large language models (LLMs) with instructions or demonstrations on many tasks has proven very effective for zero- or few-shot transfer in a variety of settings~\cite{wei2022finetuned,sanh2022multitask,instruct_gpt,min-etal-2022-metaicl,wang2022benchmarking,mishra-etal-2022-cross,flan_palm}. 
However, such instruction tuning has not been systematically explored in retrieval for several reasons.
First, large-scale instruction-annotated datasets~\cite{bach2022promptsource,wang2022benchmarking} do not include retrieval tasks.
Second, successful instruction-following LLMs are encoder-decoder or decoder-only models with tens of billions of parameters, which are difficult to be adapted for retrieval tasks, as we typically need to encode millions of documents. 

\paragraph{Retrieval with descriptions.}
{
To incorporate more fine-grained information needs, retrieval with descriptions or narratives (e.g., TREC 2004 Robust Track; \citealt{96071}) has been studied. 
Descriptions or narratives are more detailed natural language explanations that describe the information needs (i.e., desirable documents) for each query mainly for query disambiguation. 
However, early work shows that concatenating descriptions or narratives only marginally outperforms the baselines with titles only~\cite{walker1998okapi,10.1145/3404835.3462812}.
More recent work~\cite{DBLP:conf/sigir/DaiC19,10.1145/3366423.3380258} suggests that powerful encoders such as BERT~\cite{devlin-etal-2019-bert} could better incorporate rich linguistic context and boost performance.  }

\section{Task Formulation}
\label{sec:task_formulation}
{This work introduces a new task formulation, \emph{retrieval with instructions} (Figure~\ref{fig:teaser} bottom right).
}
{We are given} a large collection of $N$ documents $\mathcal{D} = \{\boldsymbol{d}_1, \ldots, \boldsymbol{d}_N\}$, a search task instruction $\boldsymbol{t}$ and a query $\boldsymbol{q}$. 
The problem of retrieval with instructions is to find a document $\boldsymbol{d} \in \mathcal{D}$ that is relevant to $\boldsymbol{q}$ according to the instruction $\boldsymbol{t}$.
Compared to the standard retrieval problem setting (e.g., Figure~\ref{fig:teaser} bottom left), the difference is the explicit definition of \emph{relevance} in the instruction~$\boldsymbol{t}$ as additional input to the system, which can be diverse and may not be fully defined by the query only~\cite{ruthven_lalmas_2003}. 
Thus, retrieval with instructions lets us build retrieval systems that are very general and task-aware---changing their relevance measure by attending to the instruction.

{
This new formulation brings both new research challenges and opportunities. For instance, a retriever is now required to modify its search behavior according to the instructions. 
On the plus side, different datasets can be naturally grouped to train a single retriever, yielding benefits from cross-task interdependence, which has been observed in instruction tuning of LLMs.
Compared to training a retriever on multiple tasks with task identifiers~\cite{maillard-etal-2021-multi}, instructions provide greater flexibility and enable zero-shot transfer via natural language instructions.
Furthermore, a multi-task instruction-following retriever obviates the need to host multiple task-specific retrievers.
}

{
As noted previously, multi-task training with instructions has not been studied in the area of retrieval due to the lack of resources and dedicated models. To facilitate the research on retrieval with instructions, we  introduce the first large-scale retrieval benchmark with expert-written annotations (Section~\ref{sec:dataaset}) and subsequently the multi-task instruction-following retrievers (Section~\ref{sec:method})}.

\section{\rib: Collections of Instruction-annotated Retrieval Tasks}
\label{sec:dataaset}
\begin{figure*}[t!]
\centering
\includegraphics[width=0.95\textwidth]{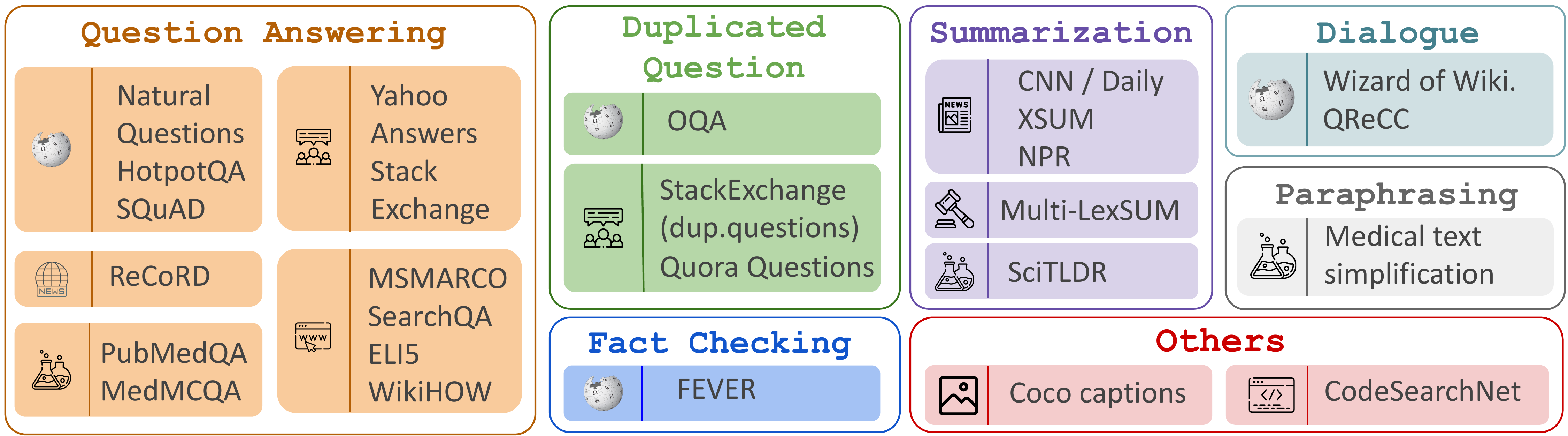}\caption{Examples of datasets included in \rib.} \label{fig:train_datasets}
\end{figure*}
To facilitate research in retrieval with natural language instructions, we build a unified large-scale retrieval dataset, \rib ({\bf B}ank of {\bf E}xplicit {\bf R}et{\bf R}ieval {\bf I}nstructions). 
{\rib consists of instructions and instances from diverse domains and tasks, by carefully curating retrieval datasets and casting diverse NLP datasets as retrieval tasks. }

\subsection{{Unified Task and Instructions Scheme }}
\paragraph{Task format.}
Each task $\mathcal{T}$ in \rib consists of a corpus $\mathcal{D}$, queries $\mathcal{Q} =\{\boldsymbol{q_1}, \ldots, \boldsymbol{q}_K\}$, and an instruction $\boldsymbol{t}$, where $K$ is the number of the queries included in the task.
An instance of each task includes a query $\boldsymbol{q}$, gold (relevant) documents $\boldsymbol{d}^{+}$, and negative (irrelevant) documents $\boldsymbol{d}^{-}$. 
For each task, an explicit intent $\boldsymbol{t}$, which can be paraphrased in multiple ways, is given. 
\paragraph{Instruction schema for retrieval.}
\label{sec:instruction_scheme}
Informative and diverse instructions are key for successful instruction tuning~\cite{sanh2022multitask}. 
{We introduce a novel scheme to define an informative instruction for retrieval tasks, which have not been studied in prior instruction-following literature. }
An instruction that sufficiently describes an arbitrary retrieval task should include: \emph{\hlc[pink]{intent}}, \emph{\hlc[cyan!30]{domain}} and \emph{\hlc[green!30]{unit}}.
{Specifically,} \emph{\hlc[pink]{intent}} describes how the retrieved text relates to the query, such as whether the text answers a question in the query or paraphrases it.
\emph{\hlc[cyan!30]{Domain}}  is the expected source or type of retrieved text, such as Wikipedia or PubMed articles. 
\emph{\hlc[green!30]{Unit}} defines the text block to retrieve, such as a sentence or a paragraph.
Table~\ref{tab:instructions} shows examples of instructions, and Appendix~\ref{sec:list_of_instructions} shows the full list.

\subsection{Dataset Collection}
\label{sec:dataset_collecitons}
\rib includes diverse datasets, including widely-used retrieval (e.g., MS MARCO;~\citealt{msmarco}), retrieval-centric datasets (NQ-open; \citealt{47761}) and non-retrieval datasets that can be repurposed as retrieval tasks. 
Figure~\ref{fig:train_datasets} shows the source datasets in \rib.  
{We manually annotate in multiple instructions per dataset}, and conduct multiple post-processing steps to provide a set of the query, gold, and negative documents. 
Table~\ref{tab:list_of_datasets} shows the full dataset list.
\paragraph{Selecting source datasets.}
We manually collect datasets from (1) KILT~\cite{petroni-etal-2021-kilt}, (2) the Sentence-Transformers Training Data for Text Embedding Models\footnote{\url{https://huggingface.co/datasets/sentence-transformers/embedding-training-data}}, and (3) manual searches in ACL anthologies and huggingface datasets\footnote{\url{https://huggingface.co/docs/datasets/index}} to cover a diverse set of tasks and domains. 
{
We observe that except for a few domains (e.g., Wikipedia) many domains do not have retrieval datasets while there are a few datasets for other NLP tasks that can be cast as retrieval (e.g., sentence paraphrase). 
Re-purposing those non-retrieval tasks as retrieval tasks enables to diversity of the domains as well as the instructions in \rib.  
}

From an initial list of more than {60} datasets, we assess whether it is suitable for repurposing as a retrieval task. Specifically, we sample 20 instances from the candidate dataset and check if the queries are self-contained.\footnote{For examples, finding a corresponding review text for the review title ``{\it I love this!}'' is under-specified.}
If the majority of queries fail this test, we exclude the corresponding dataset.
Consequently, we use 37 datasets, including more than {5 million} instances in total. 
For datasets that are orders of magnitude larger than other datasets (e.g., PAQ; \citealt{lewis-etal-2021-paq}), we randomly sample up to 300k instances, except for MS MARCO.\footnote{Prior work has shown that MS MARCO can be beneficial to many downstream retrieval tasks~\cite{izacard2022unsupervised}.}
As a result, \rib covers diverse domains (e.g., Wikipedia, scientific papers) and tasks (e.g., fact verification, dialogue response retrieval, QA). See Appendix~\ref{sec:dataset_analysis} for more details.

\begin{figure*}[t!]
\centering
\includegraphics[width=\textwidth]{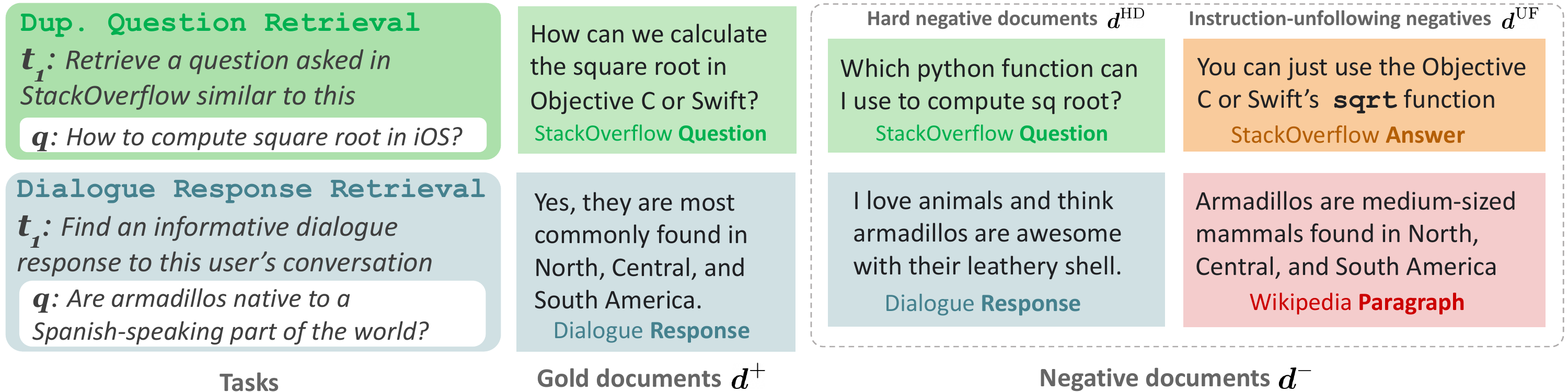}\caption{Examples of documents that are considered gold documents $\boldsymbol{d}^+$, and two types of negative documents $\boldsymbol{d}^-$: hard negatives $\boldsymbol{d}^{\rm HD}$ and instruction-unfollowing negatives $\boldsymbol{d}^{\rm UF}$ for two different query and instruction pairs. 
} \label{tab:dataset_schema}
\end{figure*}
\paragraph{{Unification and instruction annotations.}}
{
For retrieval datasets such as MS MARCO, we use the annotated gold documents as positive documents $\boldsymbol{d}^{+}$ to a given query $\boldsymbol{q}$. 
Regarding non-retrieval tasks, we use the original input sequence as a query $\boldsymbol{q}$ and the original output or given context as $\boldsymbol{d}^{+}$. 
{For instance, given a summarization dataset we use a source text and a summary as a query and a gold document, respectively. }}

For each dataset, the authors of this paper manually wrote {instructions ({8 maximum and 3.5 average number of instructions per task).}
For datasets without preprocessed retrieval targets,\footnote{For example, KILT datasets such as FEVER or NQ use the unified Wikipedia corpus. } we gather all positive and negative documents provided by the original dataset to build a single task-specific retrieval corpus $\mathcal{D}$.
{
More details are described in Appendix Section~\ref{sec:unification}. 
}

\paragraph{{Negative documents selection.}}
Negative samples are crucial for training retrieval systems~\cite{zhan-2021-optimizing}. 
{
In addition to the randomly sampled negative samples (random negative documents), we introduce two types of challenging negative samples: denoised hard negative documents $\boldsymbol{d}^{\rm HD}$ and instruction-unfollowing negative documents $\boldsymbol{d}^{\rm UF}$.}
Figure~\ref{tab:dataset_schema} shows examples of gold documents, hard negatives, and instruction-unfollowing documents. 

We mine hard negative documents $\boldsymbol{d}^{\rm HD}$ using an off-the-shelf retriever and then filter out false negative documents using an off-the-shelf reranker, following \citet{qu-etal-2021-rocketqa}. 
{In particular, we retrieve top documents from the target corpus using Contriever~\cite{izacard2022unsupervised} and then add new documents whose normalized scores predicted by a cross-encoder model, \texttt{ms-marco-MiniLM-L-12-v2}\footnote{\url{https://huggingface.co/cross-encoder/ms-marco-MiniLM-L-12-v2}} are below 0.1 as hard negative documents. }

We further introduce a new negative sampling strategy, instruction-\emph{unfollowing} negative samples $\boldsymbol{d}^{\rm UF}$, to make systems learn to \emph{follow} instructions and retrieve documents that align with the specified intents.
As shown in Figure~\ref{tab:dataset_schema}, given an instruction ``find an informative dialogue response'', a system should not retrieve a Wikipedia paragraph about armadillos. 
To obtain those instruction-unfollowing negative documents, we retrieve documents from a different task's target corpus using the same off-the-shelf Contriever and consider all those documents to be negatives since they do not satisfy the instruction.
More details about instruction-unfollowing negatives are in Appendix Section~\ref{sec:details_of_negative}. 
\section{\model: Multi-task Instructed Retriever}
\label{sec:method}
{
We now present our unified multi-task retriever \model ({\bf TA}sk-aware {\bf R}e{\bf T}riever) trained on \rib via multi-task instruction-tuning. 
}

\subsection{Model Architecture}
\label{sec:model_architecture}

\paragraph{\model-dual.}
{
\model-dual adopts a dual-encoder architecture, where a single encoder encodes queries with instructions and documents independently. 
We use maximum inner product search (MIPS) over the embeddings to find relevant documents~\cite{karpukhin-etal-2020-dense}. 
}
Formally, the similarity scores between a query $\boldsymbol{q}$ and a document $\boldsymbol{d}$, given an instruction $\boldsymbol{t}$, is calculated as follows:
\begin{equation}\label{eq:bi_enc_sim}
    {\rm s}(\boldsymbol{t}, \boldsymbol{q}, \boldsymbol{d}) = \mathbf{E}([\boldsymbol{t};\boldsymbol{q}])^T\mathbf{E}(\boldsymbol{d}),
\end{equation}
where $\mathbf{E}(\cdot)$ is the embedding function\footnote{We use a shared encoder since having separate encoders gave no additional gains in preliminary experiments.} and $[\boldsymbol{t};\boldsymbol{q}]$ is the concatenation of the instruction and query. 
For this model, document embeddings can be computed offline, improving inference efficiency at the cost of storage space~\cite{yamada-etal-2021-efficient}.

\paragraph{\model-full.}
The bi-encoder architecture is known to be less expressive since it only has limited interactions between queries and documents~\cite{colbert}, especially when the training data is limited~\cite{hofstatter2021efficiently}. 
{
To address this issue, we also explore a cross-encoder architecture~\cite{nogueira2019passage}, which computes the relevance between a query and each document by jointly encoding them with cross-attention. 
A cross-encoder model is often prohibitively expensive to scale up to millions of documents, so we first run a lightweight off-the-shelf retrieval system (e.g., a bi-encoder retrieval system) to retrieve the top documents. 
For each of these documents, our instruction-aware cross-encoder, \model-full, computes a similarity score between a query and the document $\boldsymbol{d}$ as:
}
\begin{equation}
\label{eq:ce_score}
    {\rm s}(\boldsymbol{t}, \boldsymbol{q}, \boldsymbol{d}) ={\rm FFN}(\mathbf{E}([\boldsymbol{t}; \boldsymbol{q}; \boldsymbol{d}])),
\end{equation}
where {\rm FFN} represents an additional feed-forward network that predicts whether the document follows the instruction and is related to the query. 

We explored different encoder-only and encoder-decoder models to initialize our cross-encoder; we found that initializing \model-full with encoders of T5-based instruction-following pretrained models, {namely T0-3B~\cite{sanh2022multitask} and FLAN-T5~\cite{flan_palm}, empirically leads to superior performance as found in prior work~\cite{sachan2022improving}.}
We follow the EncT5 approach~\cite{liu2021enct5} and prepended each sequence with a start-of-sequence token. 
The token representation is then fed to a newly initialized feed-forward network. 
Unlike MonoT5~\cite{nogueira-etal-2020-document}, we use their encoders only to reduce parameters and improve inference-time efficiency.

\subsection{Training \model}
\label{sec:training}
We train \model-dual and \model-full using the positive documents and {three types of negative documents
in \rib with instructions} (Figure~\ref{tab:dataset_schema}).  

\paragraph{Training \model-dual.}
During training, we combine documents annotated with the queries in \rib as well as in-batch negatives and train the model as follows: 
\begin{equation*}
    \mathcal{L} = - \log \frac{e^{s(\boldsymbol{q},\boldsymbol{d}^{+}, \boldsymbol{t})}}{\sum_{\boldsymbol{d} \in \mathcal{B}} e^{ s(\boldsymbol{q},\boldsymbol{d}, \boldsymbol{t})} },
\end{equation*}
where $\mathcal{B}$ denotes all documents in the same mini-batch~\cite{karpukhin-etal-2020-dense}.

\paragraph{Training \model-full.}
Following prior work~\cite{nogueira2019passage}, \model-full is trained with the cross entropy loss as:
\begin{equation*}
    \mathcal{L} =-\sum_{\boldsymbol{d} \in \boldsymbol{d}^{+}} \log s(\boldsymbol{t}, \boldsymbol{q}, \boldsymbol{d}) -\sum_{\boldsymbol{d} \in \boldsymbol{d}^{-}} \log (1 - s(\boldsymbol{t}, \boldsymbol{q}, \boldsymbol{d})).
\end{equation*}

\paragraph{{Knowledge distillation from \model-full to \model-dual.}}
{The default hard negatives in \rib rely on off-the-shelf models fine-tuned on MS MARCO; for some domains, the hard negative samples mined by those models can be less reliable. 
Especially for a smaller bi-encoder model, those false positive and negative samples can significantly diminish performance~\cite{qu-etal-2021-rocketqa}. 
We found that leveraging more powerful \model-full to denoise and obtain better negative and positive samples lets us distill knowledge from the powerful model to a smaller \model-dual.  
We first train \model-full on the annotated gold documents and the negative documents mined  
in \rib. }
We then re-run the denoising process described in Section~\ref{sec:dataset_collecitons} with the newly trained cross-encoder. 
Importantly, unlike the initial denoising step, we now leverage instructions and update hard negative documents $\boldsymbol{d}^{\rm HD}$ and positive documents  $\boldsymbol{d}^{+}$ based on more accurate \model-full predictions.

\begin{table*}[ht!]
\center
\small
\begin{tabular}{l  c  c  l  l l}
\toprule 
\textbf{Task}  & $|\boldsymbol{q}|$ &  $|\mathcal{C}|$ & \textbf{Domain} & \textbf{Query} & \textbf{Gold documents} \\  
\midrule
Ambig QA~\cite{min-etal-2020-ambigqa} & 1,172 & 18,809 & Wikipedia & question & duplicated question  \\
WIKIQA~\cite{yang-etal-2015-wikiqa} & 369 & 26,196 & Wikipedia & question & answer sentence \\
\midrule
SciFact~\cite{wadden-etal-2020-fact} & 300 & 5183 & Science & claim & scientific paper paragraph \\ 
\midrule
GooAQ-Technical~\cite{khashabi-etal-2021-gooaq-open} & 1,000 & 4,086 & Technical & question & StackOverflow answer \\
LinkSo-Python~\cite{linkso} & 1,000 & 485,413 &  Technical & question & StackOverflow question \\
CodeSearchNet-Python~\cite{husain2019codesearchnet} & 1,000 & 457,414 & Code & comment & Python code \\
\bottomrule
\end{tabular}
\caption{The \task evaluation. Example pairs of queries and documents are shown in Table~\ref{tab:cross_task_eval_examples}. In addition to the corpora listed above, we add the Natural Questions corpus data from BEIR~\cite{thakur2021beir}, which consists of 2,681,468 documents from Wikipedia. 
}
\label{tab:cross_task_eval}
\end{table*}

\section{Experiments}
{
We evaluate
\model on zero-shot retrieval (Section~\ref{sec:beir_lotte}) and in a challenging new evaluation setup, \task (Cross-task Cross-domain Retrieval; detailed in Section~\ref{sec:evaluatiom_benchmark}), comparing them with state-of-the-art models described in Section~\ref{sec:baselines}. 
}

\subsection{Zero-shot Retrieval Evaluations}
\label{sec:beir_lotte}
To evaluate the models' ability to perform zero-shot transfer via {\it instructions}, we run experiments on two widely used zero-shot transfer evaluation benchmarks: {\bf BEIR}~\cite{thakur2021beir} and {\bf LOTTE-pooled}~\cite{santhanam-etal-2022-colbertv2}.
{
Notably, none of the evaluation datasets and instructions overlap with \rib. 
Moreover, many tasks differ significantly from tasks used during training (e.g., argument retrieval)}.

{\bf BEIR} is a collection of diverse retrieval tasks in multiple domains (e.g., Wikipedia, biomedical) where the retrieval target is restricted to the target corpus in a single domain. We used publicly available datasets. Following \citet{promptagator}, we exclude Natural Questions, MS MARCO, HotpotQA, FEVER and CQADupStack from our evaluation targets for fair comparison since they are included either in encoders' pretraining or in \rib. 

{\bf LOTTE-Search} samples GooAQ~\cite{khashabi-etal-2021-gooaq-open} questions whose answers come from certain forums in StackExchange. 
We evaluate our model in the pooled setup, where documents come from forums in diverse domains (e.g., cooking, technical). GooAQ is not included in our training set. 
In LOTTE, our instructions specify which {\it forum domains} our system should retrieve evidence from (e.g., ``Retrieve a {\it cooking} StackExchange forum post that answers this question''). 

\paragraph{Metrics.}
Following \citet{thakur2021beir}, for BEIR, we use NDCG@10 as our primary metric on BEIR. For LOTTE-pooled, we use Success@5 ($=$ Recall@5) as our primary metric, as in the original paper~\cite{santhanam-etal-2022-colbertv2}.

\subsection{\task: Cross-task Cross-domain Retrieval Evaluation} 
\label{sec:evaluatiom_benchmark}
{
{
Normal retrieval benchmarks often assume that a system must deal with only a single intent and a closed corpus, which may oversimplify real-world scenarios: users' intents can be diverse, requiring searching in a truly open-domain environment that includes diverse documents~\cite{piktus2021web}. 
}
We introduce a more realistic evaluation setup, \task, where several retrieval tasks with different intents are pooled to form a single retrieval target containing diverse documents. This setup requires a system to adapt to a new task in a zero-shot manner and to model users' intents expressed in natural languages, to find documents aligning their expectations from an open-domain corpora. 
}

\paragraph{Tasks and queries.}
Our \task evaluation covers six datasets across three domains, namely, Wikipedia, Science, and Technical (Table~\ref{tab:cross_task_eval}) domains. 
The key challenge here includes datasets with different search intents that may not always be obvious from the queries alone. 

\paragraph{A pooled corpus.}
For the primary \emph{pooled} setup, we combine all documents from different tasks and the BEIR NQ Wikipedia corpus to form a single retrieval corpus, consisting of approximately 3.7 million documents. 
We also report the simplified \emph{closed} setup performance as an oracle setup, where a system retrieves only from the original task corpus, which is similar to BEIR. 

\paragraph{{Metrics}.}
We report NDCG@10 on both pooled and closed setups for each task. In addition, we evaluate the performance gap between the closed and pooled setups and refer to it as \emph{robustness}.
A smaller gap means that the model is distracted less by the documents from undesirable corpora.

\begin{table*}[t!]
\centering
\footnotesize
\addtolength{\tabcolsep}{-4.25pt}  
\begin{tabular}{@{}l|a|a | a| cccccccccb|c@{}}\toprule
 & \multicolumn{3}{c|}{model size \& rerank}   &\multicolumn{10}{c}{\textbf{BEIR}} & \textbf{LOTTE} \\ \midrule
& Ret. & Gen. & $K$ &TREC & NFC & FQA & ARG& TOU & DBP & SCD & CLI&SCF & Avg. & Search-Pooled \\\midrule
BM 25 & 0 & 0 & 0 & 65.6  & 32.5  & 23.6 & 31.5 & 36.7 & 31.3 & 15.8 &  21.3&  66.5 & 36.0 & 48.3  \\
Contriever & 110M & 0 & 0 & 27.4 & 31.7 & 24.5 & 37.9 & 19.3 & 29.2 & 14.9 & 15.5 & 64.9 & 29.3 & 55.5 \\ 
UPR$^\dagger$ & 3B & 0 & 0 & 60.4 & 33.3 & 45.0 & 50.3 &  21.3 & 33.8 & 17.3 & 9.5 & 69.6 & 37.8  & --  \\ 
\hline
Contriever (MS) & 110M & 0 & 0 & 59.6 &	32.8 &	32.9 &	44.6 &	23.0 & 41.3	& 16.5 & 	23.7 & 	67.7 & 	38.0 &  66.0  \\
Contriever+CE$^\dagger$& 133M & 0&  100&  70.1 &34.4 & 	36.7 &	41.3&	{ 29.8} &	47.1 &	17.1 &	25.8 &	69.2 &	41.3  & 73.5  \\
ColBERT-v2 & 110M &  0& 0 &  73.8 & 33.8 & 35.6 & 47.9 & 26.3 & 44.6 & 15.8 & 17.6 & 69.3 & 40.5 &   71.6 \\
BM25 + MonoT5 (3B)$^\dagger$ & 3B & 0& 1000& {\bf 79.6} & {\bf 38.4} & {\bf 51.2} &	28.8& 20.0 & {\bf 47.8} &	18.4 &	28.9 & 	{\bf 77.7} & 	43.4  & --   \\
GTR-11B& 4.8B & 0& 0 &  50.1 & 34.2 & 46.7 & {\bf 54.0} & 25.6 & 40.8& 16.1 & 27.0 &  63.5 & 39.8 & --  \\
SGPT-6.8B & 6.8B & 0 & 0 & 87.3 & 36.2 & 37.2 & 51.4 & 25.4 & 39.9 & {\bf 19.7} & 30.5 & 74.7 & 44.7 & --   \\
\midrule
GPL &  66M$\times$9 & 220M & 0 & 72.6 & -- & 32.8 & -- & -- &  -- & -- & -- & 66.4 & -- & --\\
Promptagator & 110M $\times$9 & 175B & 0 & 72.7 & 33.4 & 40.4 & 53.8 & 26.6 & 36.4 & 16.3 & 21.4 & 62.3 & 40.4 & --  \\
Promptagator (rank)$^\dagger$ & 220M $\times$9 & 175B & 200 & 76.0 &	36.0 & 	45.9 & 	53.1 & 	27.8  & 	41.3 & 	{ 19.1} &	22.6 & 	73.6 & 	43.9
& -- \\
\midrule
{\bf \model-dual} &110M &0 & 0 & 62.6 & 33.7 & 33.7 & 48.9 & 20.1 & 41.5 & 14.2 & 13.8 & 69.0 & 37.4 & 56.8  \\
{\bf \model-full (T0-3B)}$^\dagger$ &1.5B &0 & 100 &  71.7 & 34.0 & 42.2& 	49.8 & 	{\bf 31.2} & 	45.1 & 	17.5 & 	{ 30.0} & 	75.8 & 	{ 44.1}	& {\bf 75.7}  \\ 
{\bf \model-full (FLAN-T5)}$^\dagger$ &1.5B &0 & 100 & { 72.8} & 33.4 & 41.8 & { 51.5} & 24.9 & {46.8} & { 18.7} & {\bf  35.4} & {\bf 77.7} & {\bf 44.8}	& 73.1 \\
\bottomrule
\end{tabular}
    \caption{Zero-shot retrieval results on BEIR and LOTTE-Search (pooled). $\dagger$ indicates the models using cross-encoder-based reranking models. The first group of models use no labeled data during training. The second group uses MS MARCO at training time but have no customized task-specific data. The third group trains individual retrieval systems using automatically generated data. TREC, NFC, FQA, ARG, TOU, DBP, SCD, CLI, SCF indicates TREC-COVID~\cite{10.1145/3451964.3451965}, FIQA~\cite{10.1145/3184558.3192301}, NF Corpus~\cite{nfcorpus}, Arguana~\cite{wachsmuth-etal-2018-retrieval}, Touche-2020~\cite{touch}, DBPedia~\cite{dbpedia}, SciDocs~\cite{cohan-etal-2020-specter}, Climate- Fever~\cite{climate_fever}, and SciFact~\cite{wadden-etal-2020-fact}, respectively. ``$\times 9$'' of GPL, Promptagator means that those models train customized models for each of the datasets.  
    }
    \label{tab:beir}
\end{table*}
\begin{table*}[t!]
\centering
\footnotesize
\addtolength{\tabcolsep}{-2.pt}  
\begin{tabular}{l|cc|cc|cc|cc|cc|cc|bb| b}\toprule
& \multicolumn{2}{c}{AMB}  &
\multicolumn{2}{c}{WQA} &  \multicolumn{2}{c}{SCF}  &  \multicolumn{2}{c}{GAT} & \multicolumn{2}{c}{LSO} & \multicolumn{2}{c}{CSP} & \multicolumn{2}{b}{Avg.} &  $\Delta$ \\
&cl&pl&cl&pl&cl&pl&cl&pl&cl&pl&cl&pl&cl&pl & cl$-$pl\\
\midrule
Contriever & {\bf 96.8} & { 93.8}  & 80.9& 54.1 &  67.7 & 57.4  & 73.2 & 59.8 & { 28.0} & 26.7 & 36.7 &  36.1 & 63.9 & 54.6 & 9.3 \\
Contriever+CE & 96.6  & 47.4  & 78.2 & 58.4 & 69.1 & 61.7 &  75.4 &66.0  & {\bf 32.1} & {\bf 31.4} & 42.0 & 40.2  & 65.5 & 50.9 & 13.4 \\
\model-dual &  96.3 &  {\bf 95.3}  &  80.2 &   {\bf 63.1} & 70.1  & {66.2 } & 75.0 & 65.0 & 23.0 & 23.4 & 31.3 & 31.3 & 60.5  &  53.6 & {\bf 6.9} \\
{\bf \model-full (T0)} & 91.1 &  90.5 & { 82.1} &  52.5  &  { 74.7} & {66.2}  & {\bf 80.5} & {\bf 68.6} & 25.1 & 24.9 & { 51.4}  & {\bf 51.4} & { 67.5} & {\bf 59.1} & 8.4   \\
{\bf \model-full (FLAN)} & 94.0 &  89.6 &  8\bf 6.9 & 55.9 &  \bf 77.4 & \bf 66.3 & 78.3 & 66.7  & 18.1 & 18.4 & \bf 51.8 & 50.1 & {\bf 67.7} & 57.8 &  9.9 \\
\bottomrule
\end{tabular}
    \caption{Cross-task retrieval results.  $\Delta$ shows the performance gap between the averaged pooled performance and the averaged closed performance. AMB, WQA, SCF, GAT, LSO, CSP denote AmbigQA, WikiQA, SciFact, GooAQ-Technical, LinkSO-Python, and CodeSearchNet-Python, respectively. 
    }
    \label{tab:cross_task_results}
\end{table*}
\subsection{Baselines}
\label{sec:baselines}
We compare \model with various state-of-the-art methods. 
The first group we consider are unsupervised models that are not trained or trained only on unlabeled text;  
these include {\bf Contriever}~\cite{izacard2022unsupervised} and {\bf BM25}. 
We also compare \model with {\bf UPR}~\cite{sachan2022improving}, which reranks the Contriever results using a pretrained T0-3B. 

The second group trains retrievers and rerankers on MS MARCO or a few large-scale datasets and directly transfers them to new tasks with no adaptations, including {\bf MonoT5}~\cite{nogueira-etal-2020-document}, {\bf Contriever-MS MARCO} and {\bf Contriever-MS MARCO + Cross Encoder (CE)}, {\bf ColBERT v2}~\cite{santhanam-etal-2022-colbertv2}, {\bf GTR}~\cite{ni2021large} and {\bf SGPT-BE}~\cite{muennighoff2022sgpt}. 

The final group of models is specialized retrievers trained for each task on additional task-specific data that was automatically generated given the target corpus. 
{\bf Promptagator}~\cite{promptagator} generates large amount of in-domain data using FLAN~\cite{wei2022finetuned}, and {\bf GPL}~\cite{wang-etal-2022-gpl} generates them using DocT5Query~\cite{doc2query}. 

\subsection{Experimental Settings}
We initialize \model-full from the T0-3B encoder~\cite{sanh2022multitask} {and FLAN-T5 encoder~\cite{flan_palm}}. We sample positive and negative passages with a 1:4 ratio. 
We train \model-full up to 10k steps and take the best checkpoint based on development split performance. 
We initialize \model-dual from a Contriever-MS MARCO checkpoint~\cite{izacard2022unsupervised} and train up 30k steps. 
Per-GPU batch size is 16, and for each positive document, we sample in total 5 negative passages; 90\% of them are randomly sampled from $\mathcal{D}$, and 10\% are sampled from $\boldsymbol{d}^{\rm HD}$ and $\boldsymbol{d}^{\rm UF}$ in addition to the in-batch negative documents. 
We use 8 GPUs to train \model-full and 64 GPUs to train \model-dual.
To retrieve the initial document candidates for \model-full, we use Contriever-MS MARCO and rerank the top 100 documents.\footnote{We found that combining \model-full with the original Contreiver performs better than combining \model-full with \model-dual, possibly because \model-full uses the hard negative samples retrieved by Contriever's top-retrieved results. }
{Table~\ref{tab:list_of_instructions_beir} shows the full list of instructions for evaluations}. 
More details are in Appendix~\ref{sec:hyperparameters}. 

\section{Results}
\subsection{Zero-shot Evaluation Results}

As shown in Table~\ref{tab:beir}, \model-full significantly outperforms larger models and customized models trained on millions of synthetically generated in-domain data, advancing the state of the art on BEIR and LOTTE.  
Unlike prior methods that require additional data generation, {\model only requires a single human-written instruction to adapt to a new task. }
Compared to other methods using cross-encoder-based reranking models (e.g., BM25 + MonoT5), \model-full uses a much smaller number of paragraphs to be re-ranked, which significantly reduces latency caused by reranking at test time.

The large performance gain from Contriever-MS MARCO to \model-dual on six out of the nine BEIR tasks (e.g., SciFact, Arguana) shows the effectiveness of instructions and knowledge distillations from a larger to a smaller model.
{On the other hand, for the other three datasets (e.g., Touche-2020, Climate-FEVER), \model-dual shows large performance deterioration, degrading average performance. 
We hypothesize that model capacity (i.e., BERT-base encoder) and limited interactions between the query and document embeddings could be major bottlenecks. 
Prior work on instruction training in LLMs has shown that smaller models often do not get as much benefit as larger ones from instructions and increasing dataset size, possibly due to their limited model capacities~\cite{flan_palm,wang2022benchmarking}. 
}

For LOTTE-pooled, \model-full significantly outperforms prior state of the art by a large margin. 
We found that simply adding instructions at test time does not help, indicating that our instruction models do not simply exploit lexical matching. See more detailed results in Section~\ref{sec:robustness}.

\subsection{\task Evaluation Results}
Table~\ref{tab:cross_task_results} shows the models' \task performance. Contriever and Contriever+CE show competitive closed performance in the closed setup, as in BEIR, but they struggle in the pooled setup due to their inability to handle human instructions. 
{Especially Contriever+CE shows a large performance drop on AmbigQA-pooled by retrieving documents instead of queries due to the biases from fine-tuning on a QA dataset (i.e., MS MARCO) only.} 

\model-full shows the best closed performance and pooled performance, indicating its strong zero-shot adaptation and cross-task abilities. 
We found that a model can flexibly change its behavior based on the instructions, as shown in Table~\ref{tab:cross_task_examples}. 

{
Although the closed setting performance of \model-dual under-performs Contriever, \model-dual shows strong performance on the pooled setup, resulting in the best robustness among all. }
{This indicates that even smaller models can be guided by instructions, although they may have limited capabilities of zero-shot transfer to new tasks due to the limited model capacity and interactions.}

\section{Analysis}
We conduct a set of analyses to understand the factors that contribute to the models' ability to follow instructions, in particular, the effects of  instructions at training and inference (Section~\ref{sec:robustness}), dataset scale (Section~\ref{sec:ablation_datasets}),  model scale (Section~\ref{sec:model_scale}) and carefully-designed negative samples (Section~\ref{sec:negatives}). 
Our analysis in this section focuses on the more powerful \model-full { initialized with T0-3B.  }

\begin{figure}[t!]
\includegraphics[width=7.5cm]{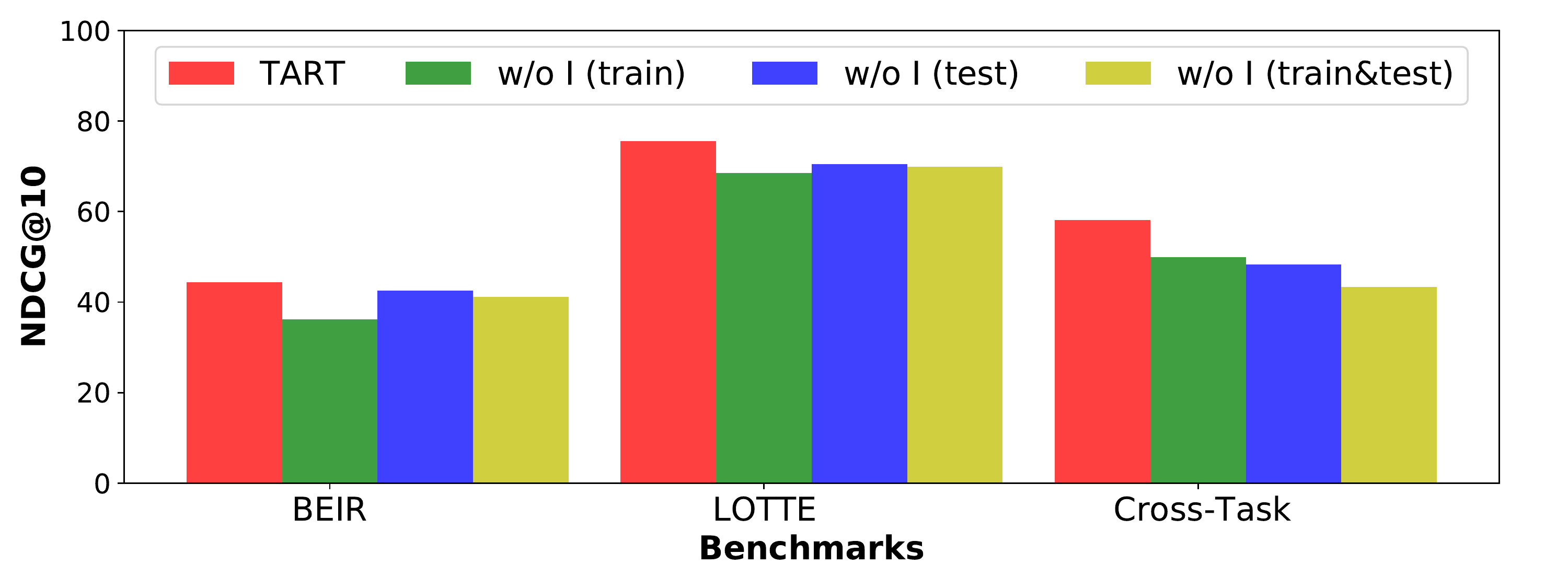}\caption{Ablations of instructions. w/o I (train), w/o I (test), and w/o (train \& test) indicate the ablations (a), (b), and (c), respectively. } \label{fig:inst_effectiveness}
\end{figure}
\begin{figure}[t!]
\centering
\begin{subfigure}[t]{.34\linewidth}
  \centering
  \includegraphics[width=0.98\textwidth,keepaspectratio]{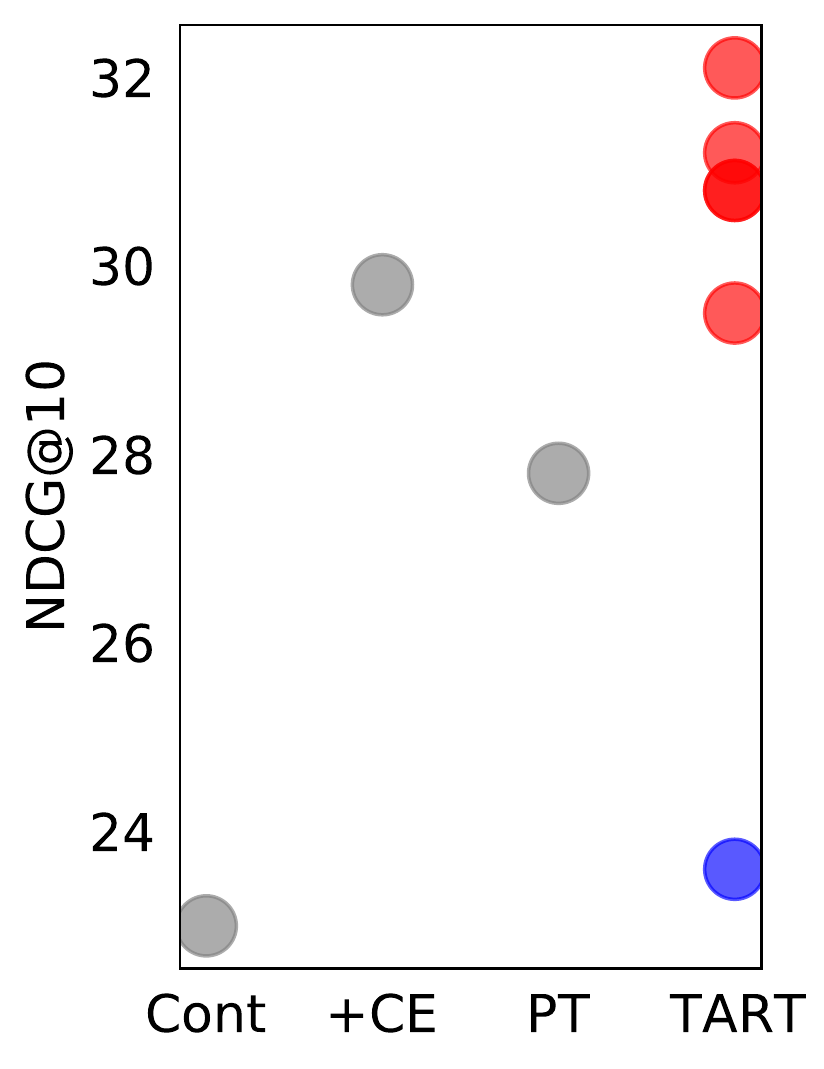}
   \captionsetup{width=0.95\textwidth}
  \caption{Touche-2020}
  \label{fig:nfcorpus}
\end{subfigure}%
\begin{subfigure}[t]{.3\linewidth}
  \centering
  \includegraphics[width=0.98\textwidth,keepaspectratio]{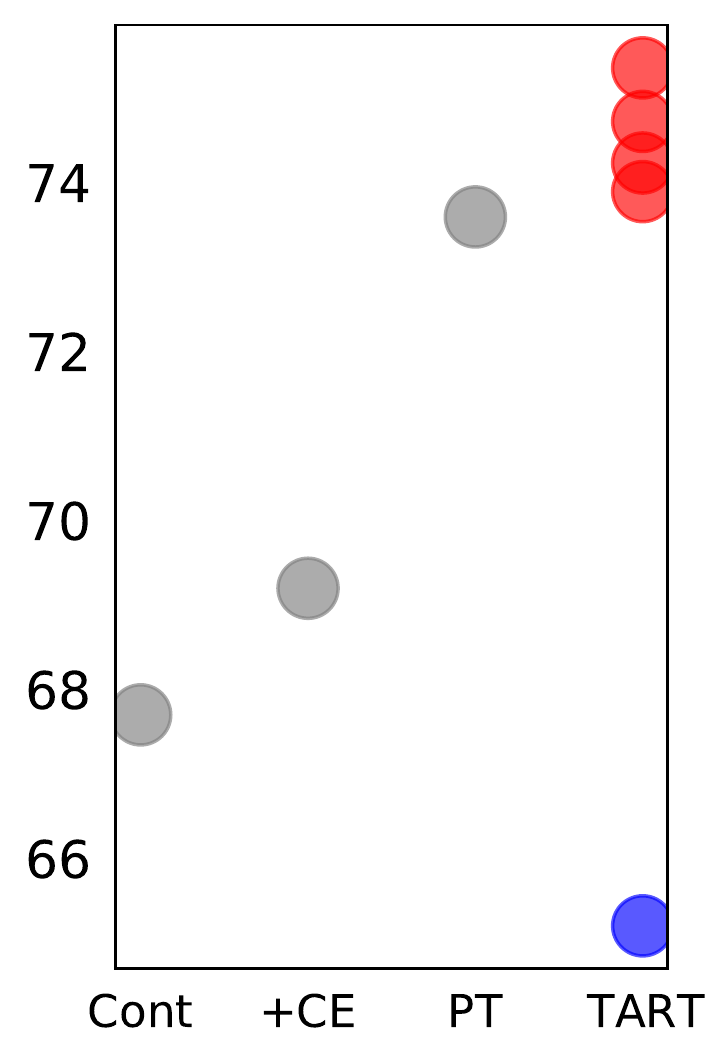}
   \captionsetup{width=0.95\textwidth}
  \caption{SciFact}
  \label{fig:scifact}
\end{subfigure}%
\begin{subfigure}[t]{.3\linewidth}
  \centering
  \includegraphics[width=0.98\textwidth,keepaspectratio]{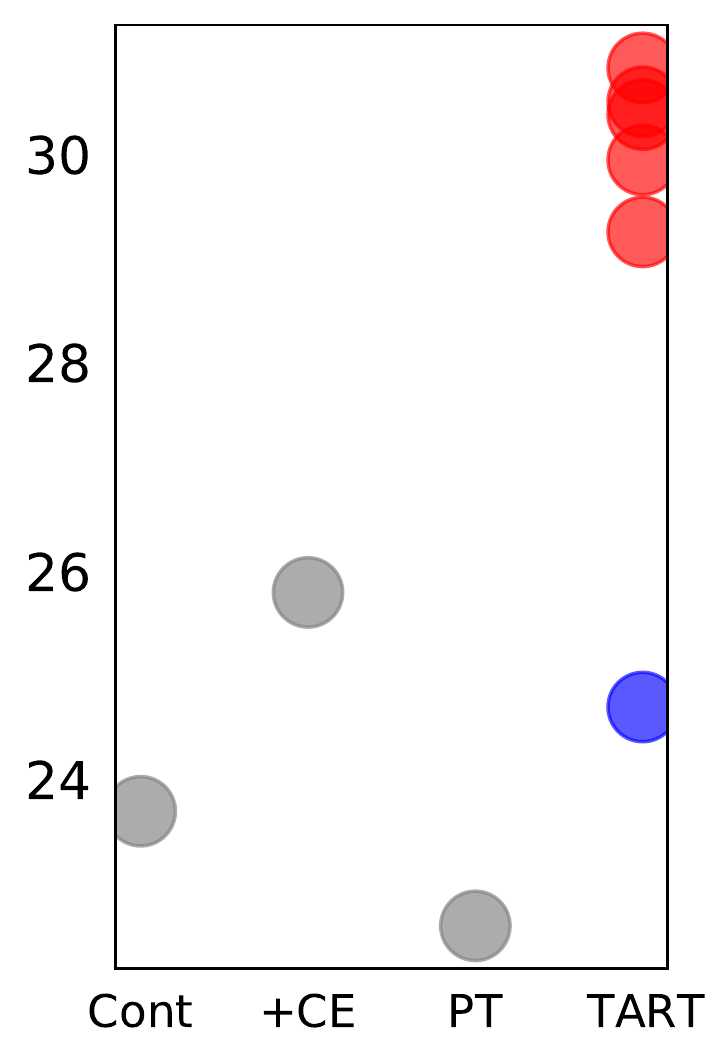} 
  \captionsetup{width=0.95\textwidth}
  \caption{C-FEVER}
  \label{fig:arguana}
\end{subfigure}%
\caption{Performance variance across different instructions. {C-Fever indicates Climate-Fever.} Blue circles indicate performance without instructions at test time, while red circles indicate performance with different instructions. 
``PT'', ``Cont'', ``+CE'' denote Promptagator, Contriever and Contriever+CE.  }
\label{fig:prompt_variance}
\end{figure}
\subsection{Effects of Instructions}
\label{sec:robustness}
\paragraph{Ablating instructions.} 
To analyze the effectiveness of instructions, we compare \model-cross with three variants: (a) {\it train without instructions, test with instructions} prepends instructions at test time only to test if the models just exploit keyword matching only at test time; (b) {\it train with instructions, test without instructions} uses \model-full without instructions at test time; (c) {\it train without instructions, test without instructions} does not use instructions at all during training and test time. 

Figure~\ref{fig:inst_effectiveness} shows the performance of those baselines. 
On all benchmarks, ablating instructions during training or test time causes a notable performance drop. 
We also see that a model trained with instructions but given no instruction at test time still yields a few performance improvements over the model trained completely without instructions, indicating the effectiveness of incorporating instructions during multi-task training. 

\paragraph{Robustness toward instructions.}
We evaluate \model-full's robustness towards diverse instructions. 
Figure~\ref{fig:prompt_variance} shows the performance variance given multiple different instructions. 
The blue circles represent the performance with no instruction.
Instructions significantly improve model performance without instructions. Although different instructions give small performance variance, \model-full often outperforms other competitive baselines when informative and correct instructions are given.
We also observe larger performance deterioration when inaccurate instructions are given. 
See Table~\ref{tab:prompt_performance_full_list} for individual instructions and performance.

 \begin{figure}[t!]
\centering
\begin{subfigure}[t]{.46\linewidth}
  \centering
  \includegraphics[width=0.9\textwidth,keepaspectratio]{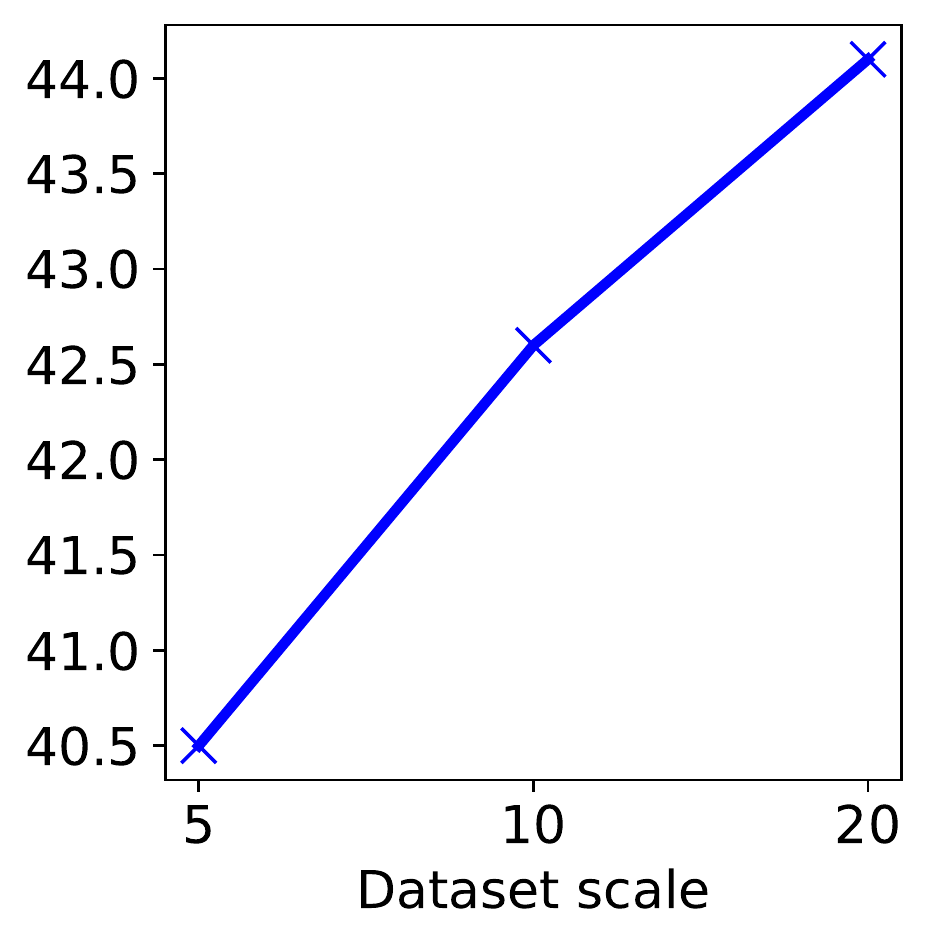}
  \captionsetup{width=0.95\textwidth}
  \caption{Dataset scale}
  \label{fig:dataset_scale}
\end{subfigure}%
\begin{subfigure}[t]{.52\linewidth}
  \centering
  \includegraphics[width=0.9\textwidth,keepaspectratio]{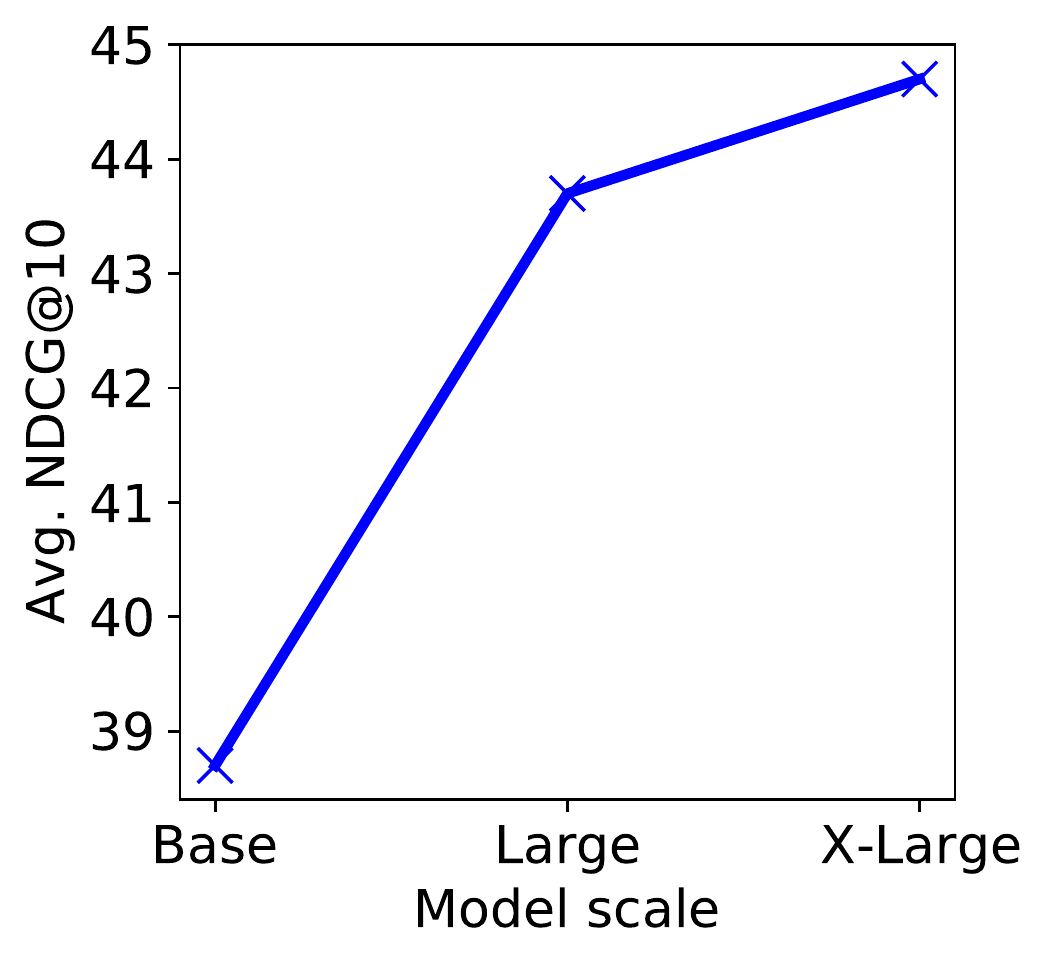}
  \captionsetup{width=0.95\textwidth}
  \caption{Model scale}
  \label{fig:model_scale}
\end{subfigure}%
\caption{Analysis of dataset and model scale. 
}
\label{fig:analysis}
\end{figure}

\subsection{Dataset Scale}
\label{sec:ablation_datasets}
To assess the effectiveness of diverse datasets, we conduct dataset ablation experiments, where we ablate datasets during training. 
Following prior work on language models instruction tuning, we conduct these experiments based on the number of datasets~\cite{wang2022benchmarking} and task clusters~\cite{wei2022finetuned}.  In addition, we run ablations based on the number of domains, where we ablate datasets based on their source domains. 
Figures~\ref{fig:dataset_scale} and \ref{fig:ablation_data} show these experiments' results. 

\paragraph{Number of datasets.}
Figure~\ref{fig:dataset_scale} shows the average performance on four BEIR datasets of \model-full trained on randomly sampled 5, 10 and 20 datasets. 
We observe that increasing the number of the training datasets helps \model to perform better. 

\paragraph{Task diversity.}
As shown in Figure~\ref{fig:ablation_data}, task diversity is a key to improve models' zero-shot transfer performance. 
QA only struggles on Arguana, where the tasks significantly differ from QA.

\paragraph{Domain diversity.}
Figure~\ref{fig:ablation_data} shows that having more diversity in training datasets' domains is also crucial, especially when the target datasets are in non-general domains. For instance, a model trained only on Wikipedia datasets struggles on Touche-2020 or SciFact, where documents come from argument websites and scientific papers, respectively. 

\begin{figure}[t!]
\includegraphics[width=8cm]{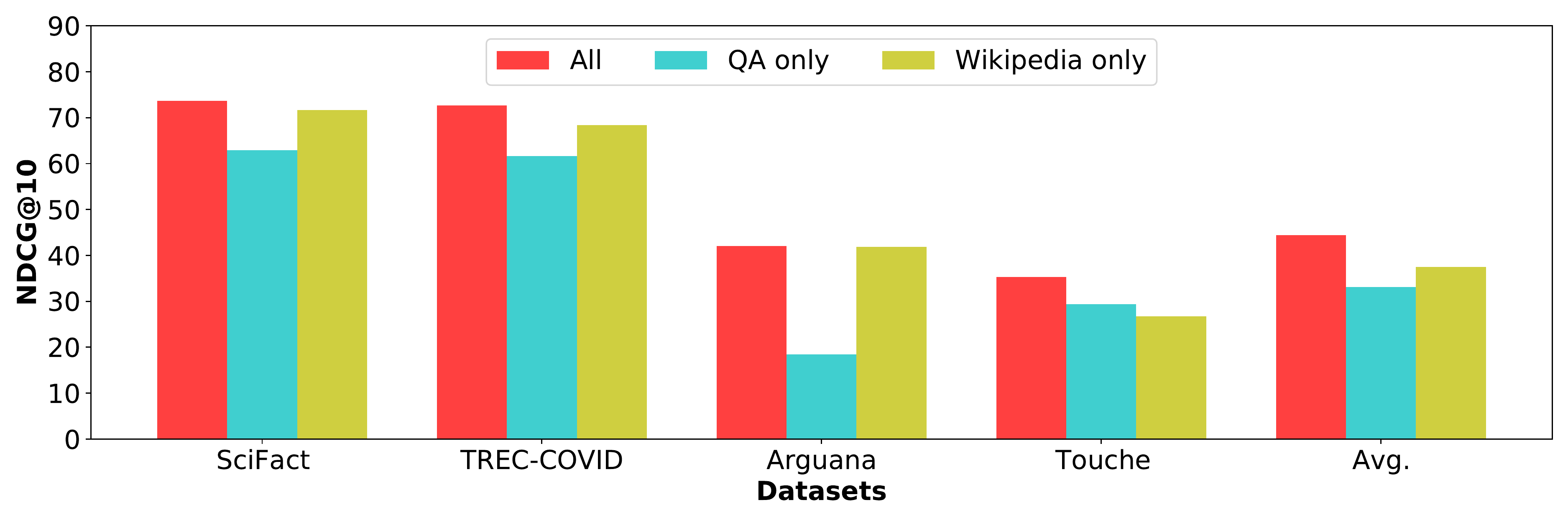}\caption{Dataset ablation results. Wikipedia-only denotes \model-full performance trained on Wikipedia-based datasets only. QA-only denotes the model trained on QA datasets only. 
} \label{fig:ablation_data}
\end{figure}

\subsection{Model Scale}
\label{sec:model_scale}
We test different \model-full sizes to see how model scale affects final performance. 
Prior work has shown that scaling up re-ranking models often improves reranking performance~\cite{rosa2022no}, and models' instruction-following abilities improve as models get larger~\cite{wang2022benchmarking,sanh2022multitask,wei2022emergent}.
We investigate how model scale affects the ability to generalize to new tasks {\it and} follow instructions. 
For a fair comparison, we train \model-full using different T5 LM-Adapt (base, large, and XL) and evaluate performance using them to rerank the top 100 Contriever results. 

Figure~\ref{fig:model_scale} shows \model-full's average performance across different model scales.  
We observe clear performance improvements by increasing model size as observed in prior work on LLM. 

\begin{figure}[t!]
\includegraphics[width=8cm]{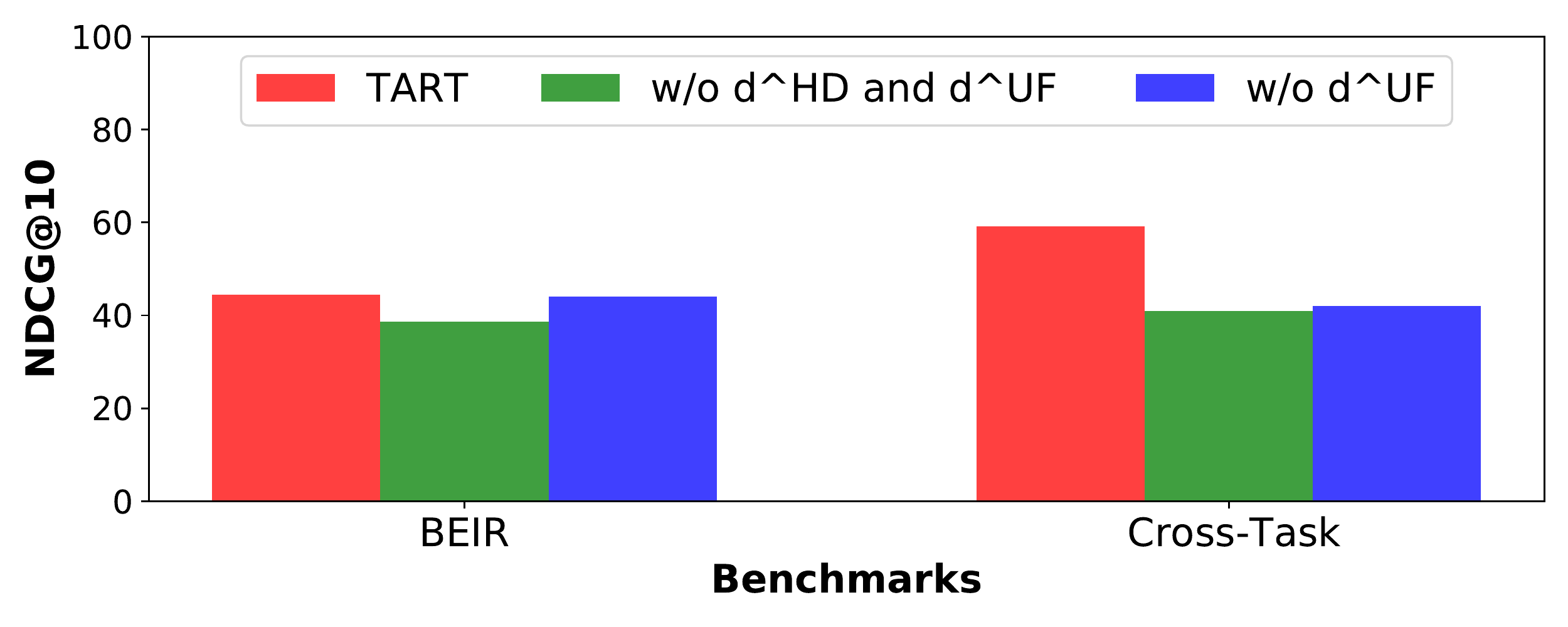}\caption{{Ablations of negative samples. 
``w/o $\boldsymbol{d}^{\rm HD}$ and $\boldsymbol{d}^{\rm UF}$'' denotes a model trained without hard and instruction-unfollowing negative documents while ``w/o $\boldsymbol{d}^{\rm UF}$''  ablates instruction-unfollowing documents only. }
} \label{fig:negative_ablations}
\end{figure}
\subsection{Negative Samples}
\label{sec:negatives}
{
We analyze the effectiveness of negative samples by ablating them during training. 
Figure~\ref{fig:negative_ablations} shows the performance of the models trained without negative samples on BEIR and \task. 
Adding more challenging negative documents (i.e.,  $\boldsymbol{d}^{\rm HD}$ and $\boldsymbol{d}^{\rm UF}$) during training largely improves the model performance on BEIR.
Moreover, the model trained without instruction-following samples (w/o $\boldsymbol{d}^{\rm UF}$) results in lower \task performance, although this model performs on par with the original \model-full on BEIR. 
This indicates that our new instruction-unfollowing negative documents largely contribute to improving the ability to distinguish instructions and are thus crucial to build a robust task-aware retrieval system. 
}

\section{Discussions and Conclusion}
{
This paper lays the foundation for building a general-purpose task-aware retriever that can follow natural language instructions. 
We introduced a new problem, \emph{retrieval with instructions}, to model users' intents explicitly.
We presented \rib, the first large-scale retrieval dataset with expert-written annotations. 
Building upon \rib, we trained the first instruction-following retrieval system by massive multi-task instruction-tuning, \model, adopting two widely used architectures. 
\model advances the state of the art on the popular zero-shot retrieval benchmarks BEIR and LOTTE as well as on our newly introduced challenging evaluation setup, \task. 
Our analysis shows that key factors to building a successful multi-task instruction-following retrieval system include informative instructions at training and test time, diversity in data and model scale, and carefully designed negative samples. 
We conclude with two interesting open questions, which future work can explore.  

{
\paragraph{Improving efficiency of instruction-following retrievers. } Although our \model-full model shows the effectiveness of instruction-tuning for retrieval, \model-dual shows some performance drop from its non-instruction-following counterpart, caused by large performance deteriorations on a few datasets. 
We hypothesize that a smaller model size (i.e., 110 million parameters) and limited interactions between query and document embeddings make building an instruction-following retrieval system more challenging. 
Future work can study the effect of scaling up bi-encoder models  as well as explore modeling architectures that enable rich interaction but are still more efficient than the cross-encoder, such as ColBERT-v2~\cite{santhanam-etal-2022-colbertv2}. 
}
}

\paragraph{Further scaling up the number of datasets.}
Retrieval tasks are excluded in prior work on instruction-following of LLMs. 
This work is the first to explore instruction-tuning in the area of retrieval, and we annotate more than 100 instructions for approximately 40 tasks, and we demonstrate the effectiveness of the dataset scale in retrieval. 
Recent work~\cite{wang2022benchmarking,flan_palm} show that scaling up the number of the training datasets improves LLMs' ability to adapt to new task via instructions. 
We open-source our instruction data and call for community efforts to collect more retrieval tasks and human-written instructions as in instruction-following for LMs~\cite{wang2022benchmarking,bach2022promptsource}, to investigate whether further increasing the number of the datasets (e.g., more than 100 datasets) improves zero-shot and cross-task retrieval.

\section*{Acknowledgements}
We thank  Allen School NLP and Meta AI researchers for their insightful discussions, and Jeff Dalton, Mike Lewis, Sheng-Chieh Lin, Sandy Kaplan and Yizhong Wang for their helpful feedback on this paper and discussions. 
\bibliography{custom}
\bibliographystyle{acl_natbib}
\newpage
\clearpage
\appendix

\section*{Appendix}
\label{sec:appendix}
\section{Further \rib Details }
\subsection{Dataset List}
Table~\ref{tab:list_of_datasets} shows all datasets we used in \rib. Table~\ref{tab:refeences} provides references for these datasets. 

\subsection{Details of Dataset Unification}
\label{sec:unification}
As shown in Table~\ref{tab:list_of_datasets}, some datasets were not originally retrieval datasets (e.g., summarization datasets). 
We describe how we convert these into the unified retrieval task format. 

\paragraph{QA.}
For QA datasets, where each instance consists of a query, a gold context, and answers, we assume the original gold context as the gold document used as a positive sample during training. 
For some exceptional datasets, we performed additional preprocessing. 
We found that ReCoRD instances are occasionally self-containing due to the nature of the cloze-style QA; therefore, for ReCoRD, we replace the original placeholder with the gold answer and use this original question with the answer as the query and the original context as a gold document. 
For MedMCQA, we use the source exam question as the query and the answer evidence as the positive document. 

\paragraph{Summarization.}
For summarization datasets, we use target summarizations as the gold document and source text as the query. 

\paragraph{Text simplifications.}
For text simplification datasets, we use source (often more complex) sentences as the query and simplified sentences as the gold document. 

\paragraph{Code search.}
We use the source comment as the query and the corresponding implication as the gold document. We exclude the python subset from \rib as we use it for \task.

\begin{table*}[ht!]
\center
\small
\begin{tabular}{l   c  c  c}
\toprule 
dataset & domain & task & unit\\\midrule
1. Altlex & Wikipedia & sentence paraphrase & sentence \\ 
2. StackExchange (title $\rightarrow$ title) & community forum & duplicated questions & title  \\ 
3. StackExchange  (query $\rightarrow$ answer) & community forum& QA & answer body  \\ 
4. Yahoo Answers (title $\rightarrow$ answers) & community forum &  QA & answer body  \\ 
5. MS MARCO & web &  QA & paragraph  \\ 
6. ELI5 & web &  QA & answer paragraph  \\ 
7. WikiHow & community forum &  QA & answer paragraph  \\ 
8. SearchQA & web &  QA & search snippets \\ 
9. AGNews & News &  summarization & news summary \\ 
10. NPR & News &  summarization & news summary \\ 
11. CodeSearchNet (java) & code & code search & Java code \\ 
12. CodeSearchNet (ruby) & code & code search & Ruby ode \\ 
13. CodeSearchNet (JavaScript) & code & code search & Java Script code \\ 
14. CodeSearchNet (Go) & code & code search & Go code \\ 
15. PAQ & Wikipedia & QA & paragraph \\ 
16. Sentence Compression & misc. & sentence compression & sentence \\
17. CNN Daily Mail & news & summarization & news summary \\
18. XSUM & news & summarization & news summary \\
19. Coco captions & image captions & caption generations & captions \\
20. Quora Duplicated Questions &  community forum & duplicated questions & questions \\
21. CCNews &  news & summarization & news summary  \\
22. FEVER  (KILT) &  Wikipedia & fact verification & paragraph  \\
23. HotpotQA  (KILT) &  Wikipedia & QA & paragraph  \\
24. NQ  (KILT) &  Wikipedia & QA & paragraph  \\
25. TriviaQA (KILT) &  Wikipedia & QA & paragraph  \\
26. WoW-KILT (knowledge) &  Wikipedia & knowledge-grounded dialogue & paragraph  \\
27. WoW-KILT (response) &  Wikipedia & knowledge-grounded dialogue & dialogue response  \\
28. medical simplification &  medical & sentence simplification & sentence  \\
29. SciTLDR &  science & summarization & paper summarization  \\
30. PubMedQA &  medical\& science & QA & abstract \\
31. MedMCQA &  medical & QA & answer explanation \\
32. Gigaword &  web & headline retrieval & headline \\
33. ReCoRD &  news &QA & news summary \\
34. MultiLexSum &  legal & summarization & legal case summary \\
35. Qrecc & Wikipedia & conversational QA & response \\
36. OQA & Wikipedia & duplicated questions & question \\
37. SQuAD & Wikipedia & QA & paragraph \\
\bottomrule
\end{tabular}
\caption{The complete list of datasets included in \rib.  Table~\ref{tab:refeences} shows references for them.
}
\label{tab:list_of_datasets}
\end{table*}
\begin{table*}[t!]
\renewcommand{\arraystretch}{1.2}
\setlength{\tabcolsep}{2pt}
\footnotesize
    \centering
    \begin{tabular}{p{15cm}}
\toprule
{\it datasets used in \rib} \\\hline
Altlex$^{*}$~\cite{hidey-mckeown-2016-identifying}, StackExchange (duplicate questions, question-title, question-question)~\cite{reimers2019sentence}, Yahoo Answers$^{*}$~\cite{yahoo_answers}, MSMARCO$^{*}$~\cite{msmarco}, ELI5$^{*}$~\cite{fan-etal-2019-eli5}, WikiHow$^{*}$~\cite{koupaee2018wikihow}, SearchQA$^{*}$~\cite{dunn2017searchqa}, AG News$^{*}$~\cite{agcorupus}, NPR$^{*}$~\cite{NPR}, CodeSearchNet$^{*}$~\cite{husain2019codesearchnet}, PAQ$^{*}$~\cite{lewis-etal-2021-paq}, Sentence Compression$^{*}$~\cite{filippova-altun-2013-overcoming}, CNN Daily Mail$^{*}$~\cite{see-etal-2017-get}, XSUM$^{*}$~\cite{narayan-etal-2018-dont}, COCO captions$^{*}$~\cite{chen2015microsoft}, Quora Duplicated Questions~\cite{quora}, CC News$^{*}$~\cite{Hamborg2017}, SQuAD$^{*}$~\cite{rajpurkar-etal-2016-squad}, FEVER$^\dagger$~\cite{thorne-etal-2018-fever}, HotpotQA$^\dagger$~\cite{yang-etal-2018-hotpotqa}, Natural Questions$^\dagger$~\cite{47761},  TriviaQA$^\dagger$~\cite{joshi-etal-2017-triviaqa}, Wizard of Wikipedia$^\dagger$~\cite{dinan2018wizard}, Medical Simplification Dataset~\cite{devaraj-etal-2021-paragraph}, SCITLDR~\cite{cachola-etal-2020-tldr}, PubMedQA~\cite{jin-etal-2019-pubmedqa}, MedMCQA~\cite{pmlr-v174-pal22a}, Gigaword~\cite{rush-etal-2015-neural}, ReCoRD~\cite{zhang2018record}, MultiLexSum~\cite{shen2022multilexsum}, Qrecc~\cite{qrecc}, OQA~\cite{Fader14}. \\\midrule
{\it datasets used during evaluations} \\\hline
TREC-COVID~\cite{10.1145/3451964.3451965}, FIQA~\cite{10.1145/3184558.3192301}, NF Corpus~\cite{nfcorpus}, Arguana~\cite{wachsmuth-etal-2018-retrieval}, Touche-2020~\cite{touch}, DBPedia~\cite{dbpedia}, SciDocs~\cite{cohan-etal-2020-specter}, Climate-Fever~\cite{climate_fever}, SciFact~\cite{wadden-etal-2020-fact}, GooAQ~\cite{khashabi-etal-2021-gooaq-open}, LinkSO~\cite{linkso}, AmbigQA~\cite{min-etal-2020-ambigqa}, WIKIQA~\cite{yang-etal-2015-wikiqa}. \\
\bottomrule
 \end{tabular}
    \caption{ References for datasets used in \rib and evaluations. We use the preprocessed versions available on the SentenceTransformers~\cite{reimers2019sentence} embedding data page~\footnote{\url{https://huggingface.co/datasets/sentence-transformers/embedding-training-data}} for the datasets with $^{*}$. We use the preprocessed versions from KILT~\cite{petroni-etal-2021-kilt} for the datasets with $^\dagger$. }\label{tab:refeences}
\end{table*}

\begin{table*}[t!]
\renewcommand{\arraystretch}{1.2}
\setlength{\tabcolsep}{2pt}
\footnotesize
    \centering
    \begin{tabular}{lp{12cm}}
\toprule
\textbf{Dataset} & \textbf{Instruction} \\\midrule
1. Altlex &  \texttt{Retrieve a sentence from Wikipedia that simplifies the following} \\
2. SE (title $\rightarrow$ title) &  \texttt{I want to find a related question asked in StackExchange. Can you find one for me?} \\
3. SE (title $\rightarrow$ title) &  \texttt{StackExchange is a community QA forum for diverse topics including technical or science. Help me to find a question body that duplicates my question} \\
4. YahooAnswers &  \texttt{Retrieve the most voted answer for this question from Yahoo Answers.} \\
5. MSMARCO &  \texttt{I want to know the answer to the question. Can you find good evidence on the web?.} \\
6. ELI5 &  \texttt{You have to answer a why / how question from users. Retrieve a Wikipedia paragraph that provides a piece of good evidence for the answer.} \\
7. WikiHow & \texttt{Find a detailed paragraph from WikiHow that explains how-to to achieve} \\
8. SearchQA &\texttt{Pick up the top web search results snippets for the following question.} \\
9. AGNews &\texttt{Find a news summary sentence corresponding to the following header.} \\
10. NPR & \texttt{Given a news article headline published at npr.org, find a corresponding summary of the news} \\
11. CodeSearchNet (Java) & \texttt{Match the following natural language instruction to Java codes} \\
12. CodeSearchNet (ruby) & \texttt{Retrieve ruby codes from GitHub commit history that implement this feature} \\
13. CodeSearchNet (JavaScript) & \texttt{Find a javascript code implementation on GitHub for the following natural language instructions} \\
14. CodeSearchNet (Go) & \texttt{Can you find a Go implementation of this?} \\
15. PAQ & \texttt{Can you answer my question by finding an article on the web?} \\
16. Sentence Compression & \texttt{You have to match this long sentence to a shorter compressed one} \\
17. CNN Daily Mail & \texttt{The following sentences are the summaries of a news article. Find the source news article.} \\
18. XSUM & \texttt{Retrieve a news article that is summarized as following.} \\
19. Coco captions & \texttt{Can you find an image caption talking about the same image as.} \\
20. Quora Dup. Questions & \texttt{Check if a Quora question is duplicated with this question.} \\
21. CC News & \texttt{I want to know the details of this news. Can you find a detailed news article on this for me?} \\
22. FEVER & \texttt{Retrieve a Wikipedia paragraph to verify this claim} \\
23. HotpotQA & \texttt{Find a paragraph that provides useful information to answer this question} \\
24. NQ & \texttt{Retrieve passages from Wikipedia to answer} \\
25. TriviaQA & \texttt{I want to find an answer for this Trivia question. Can you find some paragraphs that provide evidence from Wikipedia?} \\
26. WoW-Knowledge & \texttt{Find a Wikipedia paragraph related to the following conversation topic.} \\
27. WoW-Response & \texttt{Find a meaningful dialogue response to answer the user's question} \\
28. Medical Simplification & \texttt{Please retrieve a medical paper summary that is written in a simple language so that my patient can understand} \\
29. SciTLDR & \texttt{Find a sentence-length summary of this paper.} \\
30. PubMedQA & \texttt{Help me to find a highly related PubMed paper to answer this question.} \\
31. MedMCQA & \texttt{Find the explanation for the correct answer of this medical question.} \\
32. Gigaord & \texttt{Retrieve an extremely short summary of the following Gigaword article.} \\
33. Record & \texttt{Find a News article to verify the following sentence} \\
34. MultiLexSum & \texttt{Map this legal case summary to a sentence-long summary} \\
35. Qrecc & \texttt{You need to find a good response from a collection of previous responses and help users to know this topic more} \\
36. OQA & \texttt{Find a question that is paraphrased of this} \\
37. SQuAD & \texttt{Find a Wikipedia paragraph that answer the question} \\
\bottomrule
 \end{tabular}
    \caption{Full list of the instructions for the \rib datasets. We present one instruction per dataset. All of the instructions are available at our GitHub repository.}\label{tab:list_of_instructions}
\end{table*}

\begin{table*}[ht!]
\center
\small
\begin{tabular}{p{0.1\linewidth}  p{0.25\linewidth}  p{0.5\linewidth}}
\toprule
{\bf  Dataset }&  $\boldsymbol{q}$ & $\boldsymbol{d}^{gold}$  \\  
\hline
\multirow{3}{*}{WIKIQA} & \multirow{3}{\linewidth}{Who plays henry tudor in the white princess?} & \multirow{3}{\linewidth}{Jacob Collins-Levy as Henry VII, the King of England, Elizabeth's husband} \\
& &  \\
& &  \\\hline
\multirow{3}{*}{Ambig} & \multirow{3}{\linewidth}{Who played lead guitar for the rolling stones?} & \multirow{3}{\linewidth}{Who played lead guitar for the rolling stones since 1962?}\\
& &  \\
 & & \\\hline
\multirow{4}{*}{SciFact} & \multirow{4}{\linewidth}{The risk of male prisoners harming themselves is ten times that of female prisoners.} & \multirow{4}{\linewidth}{5-6\% of male prisoners and 20-24\% of female inmates self-harmed every year (scientific paper). } \\
& & \\
& & \\
& & \\\hline
\multirow{6}{*}{GooAQ-tech} & \multirow{6}{\linewidth}{project facet java version 1.8 is not supported eclipse mars?} & \multirow{6}{\linewidth}{You can remove and create it again, or just update it. It is because the Java version in your Project Facet is 1.8 make it 1.7. Go to Project Properties -> Project Facets and on right side checkboxes, select the java checkbox(It might be already selected) and select the version as 1.7.} \\
&&\\
&&\\
&&\\
&&\\
&&\\\hline
\multirow{3}{*}{LinkSO} & \multirow{3}{\linewidth}{could use batch normalization tensorflow} & \multirow{3}{\linewidth}{ trying implement batch normalization layer tensor flow problem running train step using tf moments get mean variance test time} \\
&&\\
&&\\\hline
\multirow{4}{*}{CodeSearch} & \multirow{4}{\linewidth}{Create a Basilisp function, setting meta and supplying a with\_meta} & \multirow{4}{\linewidth}{
def \_basilisp\_fn(f): assert not hasattr(f, "meta")  f.\_basilisp\_fn = True f.meta = None f.with\_meta = partial(\_fn\_with\_meta, f)    return f }
\\
& & \\
& & \\
& & \\
\bottomrule
\end{tabular}
\caption{\task examples data. 
}
\label{tab:cross_task_eval_examples}
\end{table*}

\begin{table*}[t!]
\renewcommand{\arraystretch}{1.2}
\setlength{\tabcolsep}{2pt}
\footnotesize
    \centering
    \begin{tabular}{lp{12cm}}
\toprule
\textbf{Dataset} & \textbf{Instruction} \\\midrule
TREC-COVID & \texttt{Retrieve Scientific paper paragraph to answer this question} \\
NF Corpus & \texttt{Retrieve Scientific paper paragraph to answer this question} \\
FIQA & \texttt{Find financial web article  paragraph to answer} \\
Arguana & \texttt{Retrieve an argument that counter argues the following paragraph} \\
Touche & \texttt{You have to retrieve an argument to this debate question}  \\
DBPedia & \texttt{Retrieve a Wikipedia introduction paragraph of the following entity}  \\
SCIDOCS & \texttt{Find scientific paper titles that are related to the following} \\
Climate-Fever & \texttt{I want to know if the following claim is true or not. Retrieve a Wikipedia paragraph on climate change for this.} \\
SciFact & \texttt{Retrieve a scientific paper sentence to verify if the following claim is true}\\\hline
WIKIQA & \texttt{Retrieve an answer sentence from Wikipedia} \\
AmbigQA & \texttt{Retrieve a question that is similar to this} \\
SciFact & \texttt{Retrieve scientific evidence to verify this claim} \\
GooAQ-technical & \texttt{Find a StackExchange forum that answers this question} \\
Codesearchnet-py & \texttt{Retrieve a python code that implements the following feature.} \\
LinkSO-Py & \texttt{You have to find a python implementation of this} \\
\bottomrule
 \end{tabular}
    \caption{Full list of the instructions used for evaluations.}\label{tab:list_of_instructions_beir}
\end{table*}

\subsection{Instructions for \rib}
\label{sec:list_of_instructions}
Table~\ref{tab:list_of_instructions} shows the full list of the instructions in \rib. 
Note that we present only one instruction for each dataset. A full list of the instructions will be released in our repository. 

\subsection{\rib Statistics}
\label{sec:dataset_analysis}
We conduct analyses on \rib to understand its domain and intent diversities.

\paragraph{Intents.}
Open-ended intents are diverse and hard to classify into fixed sets of categories. As a proxy for intents, Figure~\ref{fig:task_dist} shows the distributions of the source task categories. 
QA is the most representative category, while summarization and question duplication detection is also common due to their abundance in large-scale datasets. 
On the other hand, 
around 50 \% of the tasks do not belong to those top three categories, such as code search or caption generations, which contribute to the diversity of \rib. 
We also find that traditional non-retrieval tasks, such as sentence simplification or dialogue, can be repurposed as retrieval tasks. In Section~\ref{sec:ablation_datasets}, we analyze the effect of training data task diversity. 

\paragraph{Domains.}
Our dataset covers diverse domains. Figure~\ref{fig:domain_dist} shows that Wikipedia (e.g., NQ), web (e.g., MS MARCO), Community QA (e.g., Quora), News(e.g., CNN/Daily) dominate, while we also have some expert domains (e.g., medical, legal, technical). 
We found that although many expert domain datasets are smaller than the ones in general domains like Wikipedia, adding those high-quality expert domain datasets helps the system learn to adapt to those domains or unseen expert domains with a similar writing style (e.g., scientific papers).

\begin{figure}[t!]
\includegraphics[width=8cm]{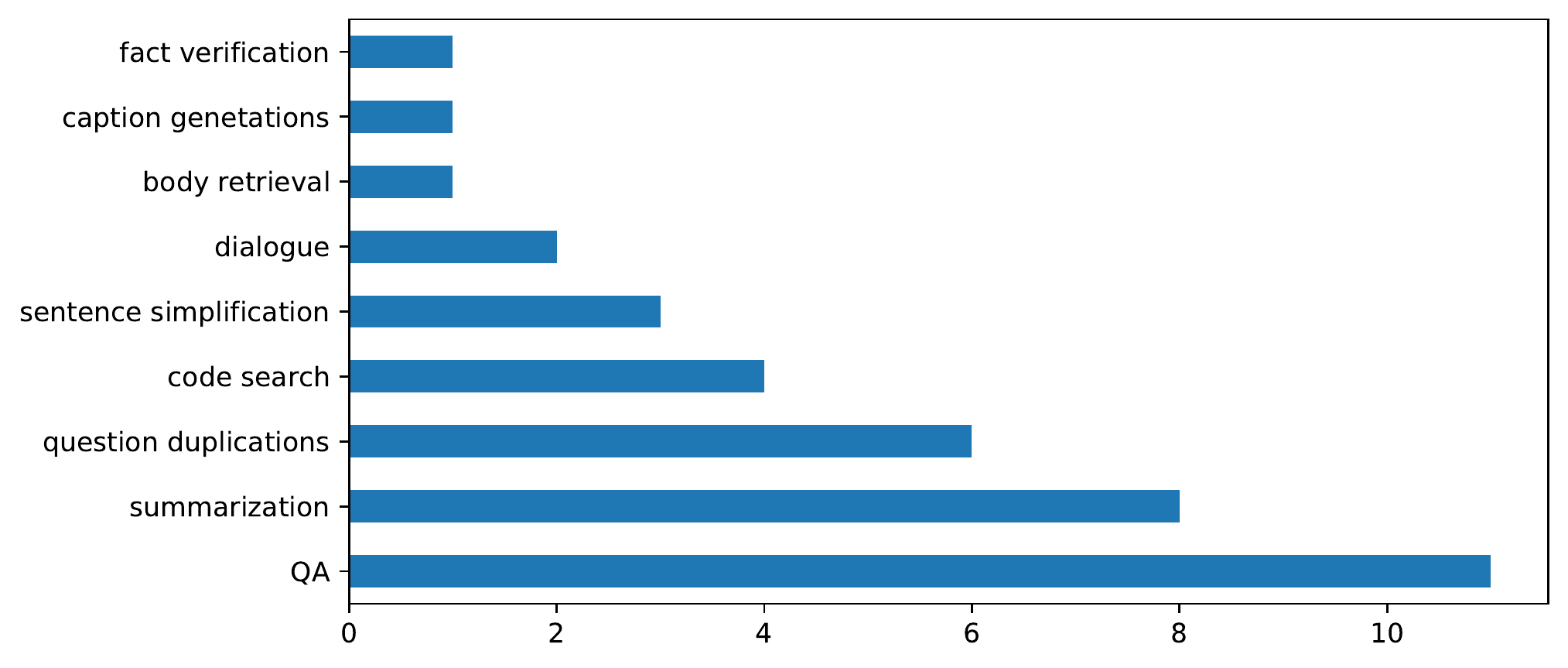}\caption{The task distributions of the datasets included in \rib. 
} \label{fig:task_dist}
\end{figure}
\begin{figure}[t!]
\includegraphics[width=8cm]{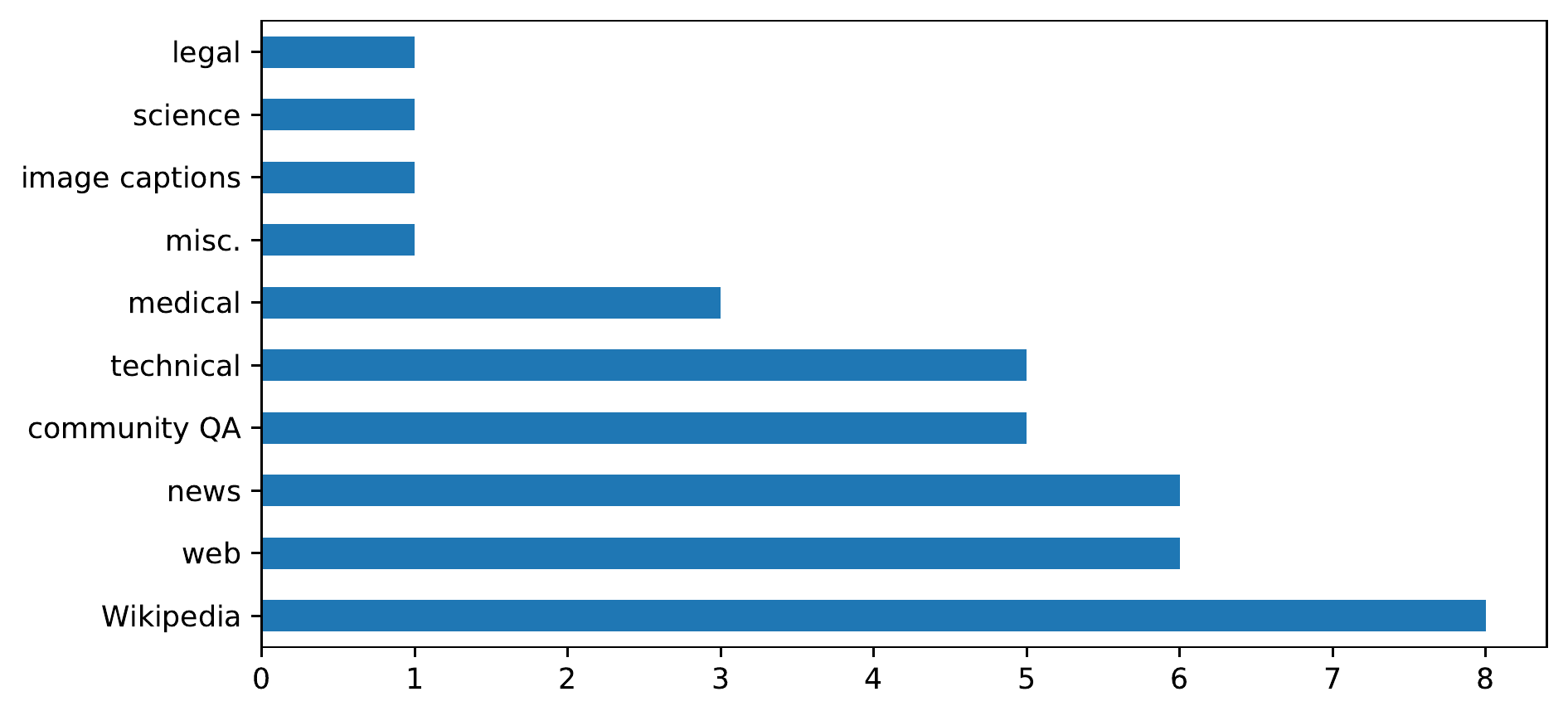}\caption{The domain distributions of the datasets included in \rib. 
} \label{fig:domain_dist}
\end{figure}

\section{Further Detail about the \task evaluation}
\paragraph{Query and corpus creations.} 
For AmbigQA, we use the official development split, including 1,172 queries, as the official test split annotations are not publicly available.  
We use all paraphrased questions for all train and development sets to form the retrieval corpus. 
For WIKIQA, we combine the development split and test split available at the huggingface datasets,\footnote{\url{https://huggingface.co/datasets/wiki_qa}} and we use the question and answer sentence pairs that are labeled as \texttt{1} as the queries for evaluations, and use the answer sentences as the gold documents.
Regarding the retrieval target, we use all sentences available in the WIKIQA dataset, including the sentences that are labeled as \texttt{0}.   
For LinkSO, we use the original datasets' test split for the python domain and sample 1,000 queries.\footnote{\url{https://sites.google.com/view/linkso}}
We find questions that are labeled as duplicated and use their corpus as our retrieval target. 
For GooAQ-technical, we sample 1,000 GooAQ questions whose answers are from \texttt{stackoverflow.com}. As 20\% of the sampled GooAQ tech queries share the same answer posts, we remove the duplicated paragraphs. 
For CodeSearchNet-Python, we use the comments describing the codes as queries and the corresponding python codes as positive documents. We sample 1,000 queries from the test split.

\paragraph{Examples.}
Examples of \task are shown in Table~\ref{tab:cross_task_eval_examples}. 
As shown, queries themselves often do not fully indicate the users' intents. 
By specifying users' intents as explicit textual instructions, our model can  effectively perform multi-task retrieval over a single pooled corpus.  
\paragraph{Human evaluations of quality. }
{To access the possibility of having false negative passages, we run an off-the-shelf retrieval system to retrieve the top 10 documents for randomly sampled 20 questions for each task, and we evaluate if any of the negative passages, especially from the non-target corpus, are indeed positive. 
We found that the false negative ratio is less than 10\%. }

\section{Modeling Details}
\subsection{Hyperparameters of \model }
\label{sec:hyperparameters}
\paragraph{\model-dual.}
We set the learning rate to be $1 \times 10^{-5}$ and warm-up steps to be 1,000. The softmax temperature is set to 0.05. 
The batch size is 1024. 
We use 7 negative samples per instance; 10\% of the time we use hard negative or instruction-unfollowing negatives, while 90\% of the time we use negative documents that are randomly sampled from the same target corpus. 
The maximum document chunk length is set to 256.

\paragraph{\model-full.}
To train a cross-encoder using the T0-3B encoder, we set the maximum sequence length to 512 and the batch size to 1, increasing the gradient accumulation steps to 8. 
We set the dropout rate to 0.1 and the learning rate to 1 $\times 10 ^ {-5}$.

\subsection{Instructions for Evaluations}
Table~\ref{tab:list_of_instructions_beir} lists the instructions used for the BEIR and \task evaluation. 

\subsection{Negative Sampling}
\label{sec:details_of_negative}
\paragraph{Mining instruction-unfollowing samples.}
To sample instruction-unfollowing samples, given a query from a target dataset, we retrieve the top 20 documents from another task's corpus using Contriever-MS MARCO. For instance, given a PubMedQA, a system should not retrieve a document from a Wikipedia paragraph. 
A list of source target task and retrieval corpus combinations is shown in  Table~\ref{tab:inst_unfollowing_list}.

\begin{table*}[t!]
\footnotesize
    \centering
    \begin{tabular}{lll}
\toprule
dataset & expected output &instruction-unfollowing corpus\\
\midrule
Gigaword & article summary &Wikipedia paragraph \\
Medical Paragraph Simplification & simplified text of medical cases &Wikipedia paragraph \\
MS MARCO & web answers & OQA questions \\
OQA & similar questions & Yahoo Answers answer \\
PubMedQA & medical paper abstract & Wikipedia paragraph \\
Qrecc & dialogue responses &Wikipedia paragraph \\
Quora & duplicated questions &Wikipedia paragraph \\
sentence compression & simplified sentence & Wikipedia paragraph \\
StackExchange (question$\rightarrow$answer) title &StackExchange answer & StackExchange title \\
StackExchange (title $\rightarrow$title) title & StackExchange title & StackExchange answer \\
Yahoo Answers & Yahoo Answers answer & Wikipedia paragraphs \\
\bottomrule
 \end{tabular}
    \caption{The list of the combinations of the dataset and corresponding {instruction-unfollowing} corpora to mine instruction-unfollowing negative documents. }\label{tab:inst_unfollowing_list}
\end{table*}

\paragraph{Sampling $\boldsymbol{d}^{-}$ for \model-full training.}
Challenging negative samples help a system to effectively learn the task. On the other hand, prior work also shows that it can lead to large performance drops in out-of-domain datasets, and having both randomly sampled negative documents and carefully designed negative documents is a key to building a system that is competitive in both in-domain and out-of-domain retrieval~\cite{ni2021large}.
To effectively combine the negative documents during training, we first combine random samples and hard negative samples, and then we randomly sample 4 negative documents per one positive document. The number of instruction-unfollowing documents, if applicable, is limited to less than 20\% of the negative documents, and we set the maximum number of instruction-unfollowing samples from certain combinations listed in Table~\ref{tab:inst_unfollowing_list} up to 10k.

\section{Further Results and Analyses}
\subsection{Qualitative Results on \task}
Table~\ref{tab:cross_task_examples} shows the qualitative examples given different instructions on \task, and Table~\ref{tab:comparison} compares \model-full with Contriever MS MARCO. 

\begin{table*}[t!]
\renewcommand{\arraystretch}{1.2}
\setlength{\tabcolsep}{2pt}
\footnotesize
    \centering
    \begin{tabular}{p{4cm}|p{11cm}}
\toprule
\multicolumn{2}{l}{\textbf{Query}: how to calculate the distance between two points using longitude and latitude} \\\hline
\textbf{Instruction} & \textbf{Top document}\\\midrule 
 Retrieve an answer post from StackOverflow to this question & SELECT getDistance(lat1,lng1,lat2,lng2) as distance FROM your\_table. Here's a MySQL function that will take two latitude longitude pairs, and give you the distance in degrees between the two points. It uses the Haversine formula to calculate the distance. \\
 Find a similar question asked in StackOverflow & tried implementing formula good two points testing yet code working distance returns. \\\toprule
 \multicolumn{2}{l}{\textbf{Query}: When did the kim family come to power? } \\\hline
\textbf{Instruction} & \textbf{Top document}\\\midrule 
find an answer sentence & Kim came to lead the Soviet-backed North's provisional government, becoming the first premier of its new government, the Democratic People's Republic of Korea (commonly known as North Korea), in 1948. He started the Korean War in 1950 with hopes to reunify the region. (Wikipedia) \\
Find a similar question & When did the kim family come to power in North Korea? (Ambig QA) \\\toprule
  \multicolumn{2}{l}{\textbf{Query}: 10\% of sudden infant death syndrome (SIDS) deaths happen in newborns aged less than 6 months } \\\hline
\textbf{Instruction} & \textbf{Top document}\\\midrule 
retrieve a scientific paper paragraph to verify this & Despite declines in prevalence during the past two decades, sudden infant death syndrome (SIDS) continues to be the leading cause of death for infants aged between 1 month and 1 year in developed countries. Behavioral risk factors identified in epidemiological studies include prone and side positions for infant sleep, smoke exposure, soft bedding, and sleep surfaces, and overheating. (Scientific paper) \\
Find a Wikipedia paragraph to verify this & By definition, SIDS deaths occur under the age of one year, with the peak incidence occurring when the infant is at 2 to 4 months of age. (Wikipedia) \\
\bottomrule
 \end{tabular}
    \caption{Examples of the model's predictions given different instructions with the same query. The queries and documents are from \task. }\label{tab:cross_task_examples}
\end{table*}

\begin{table*}[t!]
\renewcommand{\arraystretch}{1.2}
\setlength{\tabcolsep}{2pt}
\footnotesize
    \centering
    \begin{tabular}{p{7.5cm}|p{7.5cm}}
\toprule
\multicolumn{2}{l}{\textbf{Query}: 10\% of sudden infant death syndrome (SIDS) deaths happen in newborns aged less than 6 months.} \\
\multicolumn{2}{l}{\textbf{Instructions}: Retrieve a scientific paper abstract to verify this}\\\hline
\textbf{Contriever} &  \textbf{\model-full}\\
 \textcolor{red}{\xmark} By definition, SIDS deaths occur under the age of one year, with the peak incidence occurring when the infant is at 2 to 4 months of age. This is considered a critical period because the infant's ability to rouse from sleep is not yet mature (Wikipedia paragraph) &  \textcolor{green}{\checkmark} Despite declines in prevalence during the past two decades, sudden infant death syndrome (SIDS) continues to be the leading cause of death for infants aged between 1 month and 1 year in developed countries. Behavioral risk factors identified in epidemiological studies include prone and side positions for infant sleep, smoke exposure, soft bedding, and sleep surfaces, and overheating. (paper)  \\\toprule
\multicolumn{2}{l}{\textbf{Query}: Which city will host the next winter Olympics?}\\
\multicolumn{2}{l}{\textbf{Instructions}: find an answer from Wikipedia}\\\hline
\textbf{Contriever} &  \textbf{\model-full}\\
 \textcolor{red}{\xmark} Where will the next winter Olympics be held 2018? (Ambig question) & \textcolor{green}{\checkmark} The host city for the 2022 Winter Olympics, is Beijing in northern China, elected on 31 July 2015, at the 128th IOC Session in Kuala Lumpur. Beijing will be the first city ever to have hosted both the Summer and Winter Olympics. The 2022 Winter Olympics will take place between 4 and 20 February 2022. (Wikipedia paragraph) \\ \toprule
 \multicolumn{2}{l}{\textbf{Query}: use batch normalization tensorflow}\\
\multicolumn{2}{l}{\textbf{Instructions}: Can you find python code implementing this? }\\\hline
\textbf{Contriever} &  \textbf{\model-full}\\
 \textcolor{red}{\xmark} could use batch normalization tensorflow 
would like use batch normalization TensorFlow since found source code rel noreferrer core ops nn ops cc however find documented different semantics mlp cnn sure exactly bn find method called either c code copied reference  (StackOverflow post) & 
\textcolor{green}{\checkmark} 
\begin{lstlisting}
def batch_norm(inputs, training, data_format):
  outputs = tf.layers.batch_normalization(
      inputs=inputs, axis=1,
      momentum=_BATCH_NORM_DECAY, epsilon=_BATCH_NORM_EPSILON, center=True,
      scale=True, training=training, fused=True)
 return outputs
\end{lstlisting}
      (GitHub code) 
      \\\toprule
      \multicolumn{2}{l}{\textbf{Query}: how many planets is jupiter away from the sun?} \\
\multicolumn{2}{l}{\textbf{Instructions}: Can you find an answer sentence to this question for me?}\\\hline
\textbf{Contriever} &  \textbf{\model-full}\\
 \textcolor{red}{\xmark} Jupiter is the only planet whose barycenter with the Sun lies outside the volume of the Sun, though by only 7\% of the Sun's radius.[80] The average distance between Jupiter and the Sun is 778 million km (about 5.2 times the average distance between Earth and the Sun, or 5.2 AU) (Wikipedia paragraph) &  \textcolor{green}{\checkmark} Jupiter is the fifth planet from the Sun and the largest planet in the Solar System. (Wikipedia answer sentence) \\\toprule
       \multicolumn{2}{l}{\textbf{Query}: Who won the final hoh big brother 20?} \\
 \multicolumn{2}{l}{\textbf{Instructions}: a question similar to this}\\\hline
\textbf{Contriever} &  \textbf{\model-full} \\
\vspace{-0.5cm}
\begin{itemize}
\itemsep-0.2em 
    \item \textcolor{green}{\checkmark} Who won the Final HoH in the American reality show Big Brother 20? (AmbigQA)
    \item \textcolor{green}{\checkmark}  Who won the final vote in the British reality show Celebrity Big Brother 20? (AmbigQA) 
    \item  \textcolor{red}{\xmark} Caleb Reynolds was a castaway on Survivor: Kaôh Rōng; he was medically evacuated from the game, and placed 15th. Nicole Franzel returned as a HouseGuest on Big Brother 18 where she was crowned the winner and became the first female winner to win against a male in the final 2. (Wikipedia paragraph)
\end{itemize}
  &  
  \vspace{-0.5cm}
  \begin{itemize}
\itemsep-0.2em 
\item \textcolor{green}{\checkmark} Who won the final vote in the British reality show Celebrity Big Brother 20? (AmbigQA)
\item \textcolor{green}{\checkmark} Who is left in the American big brother house at the end of season 20? (AmbigQA)
\item \textcolor{green}{\checkmark} Who won the Final HoH in the American reality show Big Brother 20? (AmbigQA)
\end{itemize}
\\
\bottomrule
 \end{tabular}
    \caption{We compare \model-full outputs with the Contriever-MS MARCO~\cite{izacard2022unsupervised} predictions on \task. We show the top one prediction for the first four examples, and show the top three predictions for the bottom examples. \textcolor{green}{\checkmark} mean that the documents follow instructions while \textcolor{red}{\xmark} mean that the documents do not satisfy the instructions. }\label{tab:comparison}
\end{table*}

\subsection{Analysis of Instruction Effectiveness}
\paragraph{Full results of instruction ablations.}
Table~\ref{tab:inst_ablations_full} shows the full BEIR results of ablating instructions in Section~\ref{sec:robustness}, and Table~\ref{tab:cross_task_results_ablations} shows the ones on LOTTE and \task. 
On all of the benchmarks, removing instructions at training or test time largely hurts the performance, indicating the effectiveness of instructions. 

\begin{table*}[t!]
\centering
\small
\addtolength{\tabcolsep}{-2.25pt}  
\begin{tabular}{@{}laa | cccccccccbb}\toprule
 & \multicolumn{2}{|c|}{Using instructions}  &\multicolumn{11}{c}{\textbf{BEIR}}  \\ \midrule
 &  at training & at test &TREC & NFC & FQA & ARG& TOU & DBP & SCD & CLI&SCF & avg. & best  \\\midrule
\model-full & \checkmark &  \checkmark  &{\bf 72.8}  & 34.6 & {\bf 42.0} & {\bf 50.0} & 	{\bf 35.3} & 	46.1 & 	{\bf 18.4} & 	{ 35.2} & 	73.7 & 	{\bf 44.4}	& 5  \\\hline
 \multirow{3}{*}{Ablations}&\checkmark  &  & 61.1 & 	21.9 &  38.4 & 	39.8 & 	23.6& 36.1 & 	15.0 & 	24.7 &	65.2 &  36.2 & 0   \\
 &   &  \checkmark  & 67.6 & 	{ 34.9}  & 	40.6 & 	39.5 & 	20.5 & 	{\bf 47.1} & 	17.5 & {\bf 39.8} & 	{\bf 75.4} & 	42.5 & 3 \\
 & &  & 57.2 & 	{\bf 37.1} & 	41.3 & 	{\bf 50.0} & 	18.3 & 	41.3 & 	18.3 & 	32.5 & 	73.2 & 	41.1 &  2 \\
\bottomrule
\end{tabular}
    \caption{The full results of the instruction ablations on BEIR. TREC, NFC, FQA, ARG, TOU, DBP, SCD, CLI, SCF indicate TREC-COVID, FIQA, NF Corpus, Arguana, Touche-2020, DBPedia, SciDocs, Climate-Fever, and SciFact, respectively. 
    }
    \label{tab:inst_ablations_full}
\end{table*}

\begin{table*}[t!]
\centering
\footnotesize
\addtolength{\tabcolsep}{-2.5pt}  
\begin{tabular}{laa|c|ccccccb}\toprule
 & \multicolumn{2}{|c|}{Using instructions} & LOTTE & \multicolumn{7}{c}{\task} \\
& at training & at test   & & AMB & WQA & SCF & GAT & LSO & CSP & avg.  \\
\midrule
\model-full & \checkmark & \checkmark & {\bf 75.7} & {\bf 90.5} & 52.5 & {\bf 66.2} & {\bf 68.6} & 24.9 & {\bf 51.4} & {\bf 59.1} \\\hline
 \multirow{3}{*}{Ablations}& \checkmark & & 68.5 & 59.3 & {\bf 54.4} & 61.7 & 62.0 & 15.1 & 46.8 & 49.9 \\
  & & \checkmark & 70.5 & 40.1 & 47.2 & 64.0 & 69.5 & {\bf 25.5} &  43.7 & 48.3 \\
  & &  & 69.9 & 34.5 & 32.5 & 60.8 & 58.2 & 24.2 & 49.3 & 43.3  \\
\bottomrule
\end{tabular}
    \caption{: Instruction ablations on LOTTE (Search pooled) and \task (pooled) evaluation. AMB,
WQA, SCF, GAT, LSO, CSP denotes AmbigQA, WikiQA, SciFact, GooAQ-Technical, LinkSO-Python, and CodeSearchNet-Python, respectively.  
    }
    \label{tab:cross_task_results_ablations}
\end{table*}

\paragraph{Examples of prompts with performance.}
Table~\ref{tab:prompt_performance_full_list} shows the instructions and \model-full performance on three BEIR datasets.  
We also provide a comparison of the model performance when uninformative instructions are given in Table~\ref{tab:prompt_performance}. 
We see that more informative and related instructions often result in a strong performance, while irrelevant instructions degrade it. 

\begin{table*}[t!]
\renewcommand{\arraystretch}{1.2}
\setlength{\tabcolsep}{2pt}
\footnotesize
    \centering
    \begin{tabular}{lp{9cm}r}
\toprule
\textbf{Dataset} & \textbf{Instruction} & \textbf{NDCG@10}\\\midrule
\multirow{7}{*}{SciFact} &  \texttt{Find a scientific paper sentence to verify this questions} & 75.4 \\
 & \texttt{Retrieve a scientific paper abstract to verify this claim} & 75.7\\ 
  &\texttt{can you retrieve reliable scientific evidence to check if the following claim is true or not?} &  74.3 \\
  &  \texttt{please retrieve evidence for me to verify the following} &  73.8 \\
   &     \texttt{a scientific paper sentence supporting or refuting the following statement} &  74.7 \\
    \hline
\multirow{5}{*}{Touche-2020} &\texttt{retrieve an argument paragraph to answer this question} & 30.6 \\
&\texttt{retrieve a paragraph to answer this debate question} & 30.9 \\
&\texttt{Find a opinion to this debate question} & 29.5 \\
&\texttt{retrieve an argument paragraph that supports this debate question to this debate question} & 31.2 \\\hline
\multirow{8}{*}{Climate-FEVER} &  \texttt{Retrieve a scientific paper abstract to verify the following claim} &29.3 \\
& \texttt{Retrieve a Wikipedia paragraph to answer this question} & 30.4 \\ 
 & \texttt{Retrieve a Wikipedia paragraph to verify the following claim about climate change} &  30.8 \\
 &   \texttt{I want to know if the following claim is true or not. Can you find Wikipedia evidence?} & 30.6 \\
  &      \texttt{Find a Wikipedia paragraph to verify the following claim} & 30.8 \\
\bottomrule
 \end{tabular}
    \caption{Performance on SciFact, Climate-FEVER and Touche-2020 with different instructions. }\label{tab:prompt_performance_full_list}
\end{table*}

\begin{table*}[t!]
\renewcommand{\arraystretch}{1.2}
\setlength{\tabcolsep}{2pt}
\footnotesize
    \centering
    \begin{tabular}{lp{10cm}r}
\toprule
\textbf{Dataset} & \textbf{Instruction} & \textbf{NDCG@10}\\\midrule
\multirow{3}{*}{SciFact} & \textcolor{green}{\checkmark} \texttt{Retrieve a scientific paper abstract to verify this claim} & 75.7 \\
 & \textcolor{red}{\xmark}  \texttt{Retrieve a {Wikipedia paragraph} to verify the following claim} & 74.0 \\
 & \texttt{[NULL]} & 69.1 \\\hline
\multirow{3}{*}{Arguana} & \textcolor{green}{\checkmark} \texttt{Retrieve an article that contradict the following paragraph} & 50.6 \\
 & \textcolor{red}{\xmark} \texttt{Retrieve {a Wikipedia paragraph that answers this question}} & 47.3 \\
 & \texttt{[NULL]} & 39.8 \\\hline
\multirow{3}{*}{Touche-2020} & \textcolor{green}{\checkmark} \texttt{Retrieve an argument for this topic} &29.6 \\
 & \textcolor{red}{\xmark} \texttt{retrieve a Wikipedia passage that answers this question} &26.7 \\
 & \texttt{[NULL]} & 22.1 \\
\bottomrule
 \end{tabular}
    \caption{Full list of the instructions used for evaluations. \texttt{[NULL]} means that at inference time, no instruction is given to \model-full. \textcolor{green}{\checkmark} means a correct instruction, while  \textcolor{red}{\xmark}  means incorrect instructions. }\label{tab:prompt_performance}
\end{table*}

\subsection{Analysis on Model and Dataset Scale}
Table~\ref{tab:beir_scale} shows the NDCG@10 across different model scales. 
We compare the \model-full initialized with different sizes of T5-LM-adapt for a fair comparison. 
We see in general that larger models perform better. 

Table~\ref{tab:dataset_scale_beir} shows the full BEIR results of \model-full trained on varying numbers of datasets. 
We see that as we increase the number of datasets used during training, model performance often improves, which is consistent with previous work on instruction-tuning in LLMs~\cite{wang2022benchmarking}.

\begin{table*}[t!]
\centering
\small
\addtolength{\tabcolsep}{-2.25pt}  
\begin{tabular}{@{}l|a| cccccccccbb}\toprule
 & model size   &\multicolumn{11}{c}{\textbf{BEIR}}  \\ \midrule
 pretrained models &&TREC & NFC & FQA & ARG& TOU & DBP & SCD & CLI&SCF & avg. & best  \\\midrule
T5-LM-base&110M & 62.9 & 29.7 & 33.9 & 	37.8 & 30.8 & 38.6 & 15.1& 	29.2 & 	70.7 &  38.7 & 0\\
T5-LM-large & 385M & {\bf 73.3 }& {\bf 34.2} & 40.2 & {\bf 47.1} & 32.8 & 45.3 & 18.2 & 35.2 & 74.9 & 43.7	& 3  \\
T5-LM-XL &1.5B & 71.6 & 33.1 & {\bf 41.8} & 43.1 & {\bf 34.0} & {\bf 46.0 }& {\bf 18.5} & {\bf 38.3} & {\bf 75.5} & {\bf 44.7} & {\bf 6}	 \\
\bottomrule
\end{tabular}
    \caption{Zero-shot retrieval results for different sizes of \model-full on BEIR. TREC, NFC, FQA, ARG, TOU, SCD, CLI, SCF indicate TREC-COVID, FIQA, NF Corpus, Arguana, Touche-2020, DBPedia, SciDocs, Climate-Fever, and SciFact, respectively. 
    }
    \label{tab:beir_scale}
\end{table*}
\begin{table*}[t!]
\centering
\small
\addtolength{\tabcolsep}{-2.25pt}  
\begin{tabular}{@{}l|a| cccccccccbb}\toprule
 & dataset number   &\multicolumn{11}{c}{\textbf{BEIR}}  \\ \midrule
  pretrained models& &TREC & NFC & FQA & ARG& TOU & DBP & SCD & CLI&SCF & avg. & best  \\\midrule
T5-LM-XL &5 & 63.3 & 28.3 & 37.6 & 	47.8 & 24.3 & 42.3 & 17.0 & 30.8 & 	73.4 &   40.5  & 0 \\
T5-LM-XL & 10 & 68.8 & 30.5 & 39.5 & 47.5 & 29.4 & {\bf 46.7} & {\bf 18.2} &  26.9 & {\bf 76.0} &  42.6& 3 \\
T5-LM-XL &20 & 	{\bf 71.0 }& {\bf  33.7} & {\bf 41.7} & {\bf 48.7} & {\bf 33.2} & 46.1 & {\bf 18.2} & {\bf 29.8} & 74.7 & {\bf 44.1} & {\bf 6 } \\
\bottomrule
\end{tabular}
    \caption{Zero-shot retrieval results for different training dataset scales of \model-full on BEIR. TREC, NFC, FQA, ARG, TOU, SCD, CLI, SCF indicate TREC-COVID, FIQA, NF Corpus, Arguana, Touche-2020, DBPedia, SciDocs, Climate-Fever, and SciFact, respectively. 
    }
    \label{tab:dataset_scale_beir}
\end{table*}

\subsection{Analysis on Different Pre-trained Models}
Our \model-full is initialized with the T0-3B encoder. 
We experiment with more recent pretrained instruction-following models: FLAN-T5-XL~\cite{flan_palm} and Tk-Instruct~\cite{wang2022benchmarking}, which are trained on the order of magnitude of more datasets. 
We analyze \model-full performance when we initialize encoders using different pre-trained encoder models, including the ones that are released recently. 
Table~\ref{tab:diff_enc} shows the results of \model-full, when the encoder is initialized with three different recent instruction-following pretrained models, T0-3B, FLAN-T5-XL~\cite{flan_palm} and Tk-Instruct-3B~\cite{wang2022benchmarking}. 
FLAN-T5 shows the best average BEIR performance, outperforming TART-full by 0.7 NDCG@10. 
Tk-Instruct shows a notable performance drop on some datasets (e.g., TREC COVID), resulting in slightly lower performance than the original \model-full (T0-3B). 

\begin{table*}[t!]
\centering
\small
\addtolength{\tabcolsep}{-2.25pt}  
\begin{tabular}{@{}l|cccccccccb}\toprule
  &\multicolumn{10}{c}{\textbf{BEIR}}  \\ \midrule
 pretrained models &TREC & NFC & FQA & ARG& TOU & DBP & SCD & CLI&SCF & avg.   \\\midrule
T0-3B  & 71.7 & {\bf 34.0} & {\bf 42.2} & 	49.8 & 	{\bf 31.2} & 	45.1 & 	17.5 & 	{30.0} & 	75.8 & 	{ 44.1} \\
FLAN-T5 & { 72.8} & 33.4 & 41.8 & { 51.5} & 24.9 & {46.8} & { 18.7} & {\bf  35.4} & {\bf 77.7} & {\bf 44.8}		 \\
Tk-Instruct &  65.4 & 34.7 & 32.3 & 44.5 & 24.3 & 42.3 & 19.2 & 34.0 & 76.2 & 41.4 \\
\bottomrule
\end{tabular}
    \caption{Zero-shot retrieval results for \model-full initialized using different pretrained models on BEIR. TREC, NFC, FQA, ARG, TOU, SCD, CLI, SCF indicate TREC-COVID, FIQA, NF Corpus, Arguana, Touche-2020, DBPedia, SciDocs, Climate-Fever, and SciFact, respectively. 
    }
    \label{tab:diff_enc}
\end{table*}

\end{document}